\documentclass[lettersize,journal]{IEEEtran}
\usepackage{amsmath,amsfonts}
\usepackage{algorithmic}
\usepackage{algorithm}
\usepackage{array}
\usepackage[caption=false,font=normalsize,labelfont=sf,textfont=sf]{subfig}
\usepackage{textcomp}
\usepackage{stfloats}
\usepackage{url}
\usepackage{verbatim}
\usepackage{graphicx}
\usepackage{cite}
\usepackage{arydshln}
\usepackage{subfloat}
\usepackage[table]{xcolor}
\usepackage{caption}
\usepackage{multirow}
\usepackage{bm}
\newcommand{\etal}{{et al}. }
\newcommand{\ie}{{i}.{e}., }
\newcommand{\eg}{{e}.{g}., }
\newcommand{\etc}{{etc}.}
\newcommand{\tabincell}[2]{\begin{tabular}{@{}#1@{}}#2\end{tabular}}

\begin{document}

\title{Exploring Dynamic Transformer for Efficient Object Tracking}

\author{Jiawen Zhu, Xin Chen, Haiwen Diao, Shuai Li, Jun-Yan He, \\
	Chenyang Li, Bin Luo, Dong Wang, Huchuan Lu
	
	\thanks{
		Jiawen Zhu, Xin Chen, Haiwen Diao, Dong Wang, and Huchuan Lu are with the Dalian University of Technology, Dalian 116024, China.  
		(E-mail: jiawen@mail.dlut.edu.cn; chenxin3131@mail.dlut.edu.cn;
		diaohw@mail.dlut.edu.cn; wdice@dlut.edu.cn; lhchuan@dlut.edu.cn). 
		
		Shuai Li is with the Hong Kong Polytechnic University, Hong Kong. (E-mail: csshuaili@comp.polyu.edu.hk). 
		
		Jun-Yan He, Chenyang Li, and Bin Luo are with Tongyi Lab, Alibaba Group, Shenzhen 518000, China. (E-mail: junyanhe1989@gmail.com; lichenyang.scut@foxmail.com; luwu.lb@alibaba-inc.com).
	}
}

\markboth{Journal of \LaTeX\ Class Files,~Vol.~14, No.~8, AUGUST~2021}%
{Exploring Dynamic Transformer for Efficient Object Tracking}

\maketitle

\begin{abstract}
	The speed-precision trade-off is a critical problem in visual object tracking, as it typically requires low latency and is deployed on resource-constrained platforms. 
	Existing solutions for efficient tracking primarily focus on lightweight backbones or modules, which, however, come at a sacrifice in precision.
	In this paper, inspired by dynamic network routing, we propose DyTrack, a dynamic transformer framework for efficient tracking.
	Real-world tracking scenarios exhibit varying levels of complexity.
	We argue that a simple network is sufficient for easy video frames, 
	while more computational resources should be assigned to difficult ones.
	DyTrack automatically learns to configure proper reasoning routes for different inputs, thereby improving the utilization of the available computational budget and achieving higher performance at the same running speed.
	We formulate instance-specific tracking as a sequential decision problem and incorporate terminating branches to intermediate layers of the model.
	Furthermore, we propose a feature recycling mechanism to maximize computational efficiency by reusing the outputs of predecessors. 
	Additionally, a target-aware self-distillation strategy is designed 
	to enhance the discriminating capabilities of early-stage predictions by mimicking 
	the representation patterns of the deep model. 
	Extensive experiments demonstrate that DyTrack achieves promising speed-precision trade-offs with only a single model. For instance, DyTrack obtains 64.9\% AUC on LaSOT with a speed of 256 $fps$.
\end{abstract}


\section{Introduction}
\label{sec:intro}

\IEEEPARstart{V}{isual}
object tracking aims to continuously locate an arbitrary target object in a video sequence.
The impressive performance of recent tracking algorithms~\cite{siameserpn,yu2022learning,transt,srrt,ostrack,mixformer} is largely attributed to sophisticated network architectures~\cite{alexnet, resnet,vit}.
In particular, the successful integration of the transformer~\cite{attention_is_all} has significantly advanced tracking performance, setting new state-of-the-arts.
Despite top-notch performance, 
these trackers come at the expense of increased latency cost and, therefore, have limitations such as deployment on resource-constrained devices.
They fall into “Cannikin's Law” where higher burden structures are designed to handle a small number of complex cases.
While in real-world object tracking, efficiency and performance should be both important.

Some studies~\cite{eco, atom, lighttrack, fear, asymtrack} dedicated to improving tracking efficiency have emerged. ECO~\cite{eco} proposes a factorized convolution operator with lower linear complexity to reduce the number of parameters in DCF~\cite{dcf} model. 
LightTrack~\cite{lighttrack} utilizes the power of neural architecture search (NAS) to find lightweight network modules (\ie backbone and head). 
These modules form a lightweight tracker, with fewer floating point operations per second (FLOPs) and parameters.
Recently, Borsuk~\etal~\cite{fear} present a fast siamese tracker consisting of a compact feature extraction block and a fusion module. 
These methods achieve satisfactory efficiency by dedicated designing or searching lightweight architectures or modules. 
However, suffering from the “speed or precision dilemma”, these efficient trackers lean towards speed, thus partially lacking the capabilities to handle complex and corner cases. 

\begin{figure}[t]
	\centering
	\includegraphics[width=0.99\linewidth]{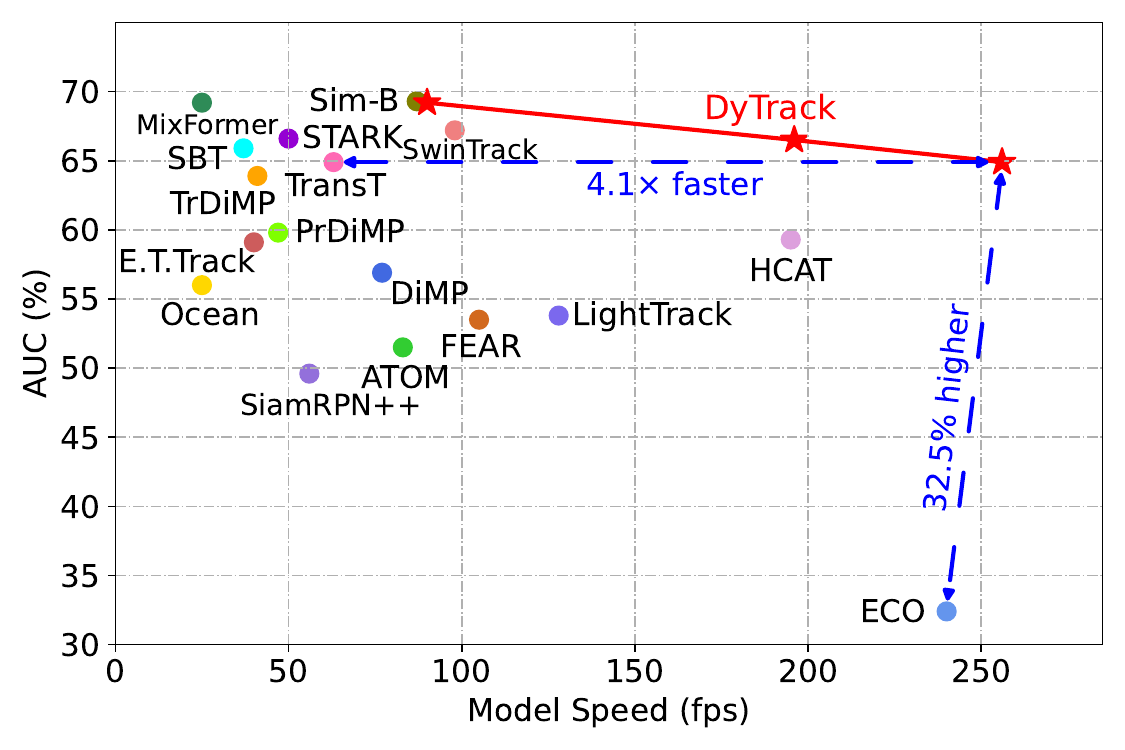}
	\vspace{-3mm}
	\caption{Speed-precision trade-off comparison on LaSOT~\cite{lasot} benchmark.
		The top right points are in the Pareto front.
		DyTrack performs better than other competing trackers at  similar running speeds (\eg 32.5 points higher than ECO~\cite{eco}), and runs faster than others when attaching the same performance (\eg 4.1 times faster than TransT~\cite{transt}). 
	}
	\label{fig:speed_vs_performance}
	\vspace{-4mm}
\end{figure}

\begin{figure}[t]
	\centering
	\begin{subfloat}
		\centering   
		\includegraphics[width=0.4123\linewidth]{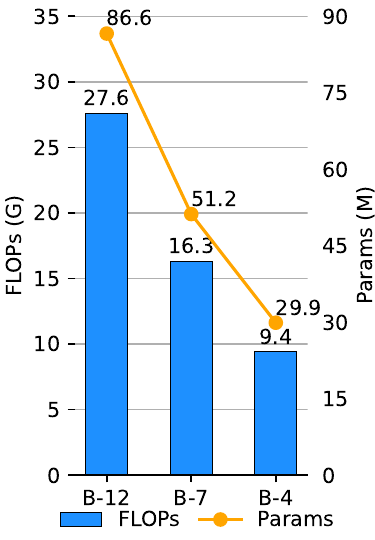}
	\end{subfloat}  
	\hspace{2mm}
	\begin{subfloat}
		\centering   
		\includegraphics[width=0.4123\linewidth]{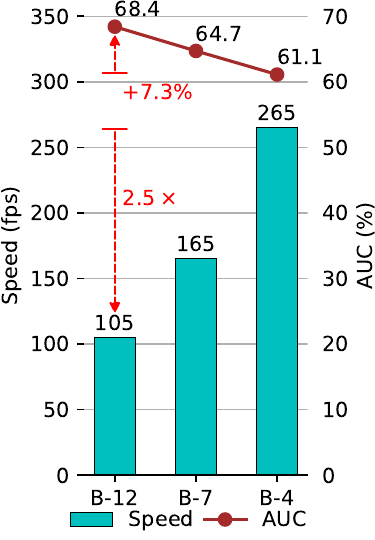}
	\end{subfloat}
	\vspace{-0mm}
	\caption{Comparison between models with various depths.}
	\vspace{-1mm}
	\label{fig:motivation}
\end{figure}

A natural question should be: \textit{is there an approach that can achieve a satisfactory trade-off between speed and precision?}
It is a realistic and challenging task to develop a tracker that can meet the accuracy and speed requirements at the same time.
We find that previous works, whether focusing on performance or efficiency, all follow the “one-size-fits-all” paradigm. 
They leave the static computational structures to learn feature propagation equally to handle all tracking scenarios, which may not be optimal.
In fact, different inputs have their distinct complexities~\cite{pmn}.
Real-world tracking scenarios have considerable variations between sequences and frames (\eg motion states, backgrounds, distractors, \etc).
Different inputs could be assigned specialized network structures or propagation routines with respect to their difficulties.
For a pilot analysis, we train baseline models with different numbers of stacked transformer layers, the precision, FLOPs, parameters, and speed results are reported in Figure~\ref{fig:motivation}.
We can observe that when the model depth reduces from 12 to 4 layers (B-12 to B-4), the inference speed increases from 105$fps$ to 265$fps$ (2.5x faster), while the AUC (area-under-the-curve) 
score only decreases by 7.3\% (68.4\% to 61.1\%). 
It indicates that computations are wasted on most easy scenarios where a simple network is sufficient.

\begin{figure}[t]
	\centering
	\includegraphics[width=0.48\textwidth]{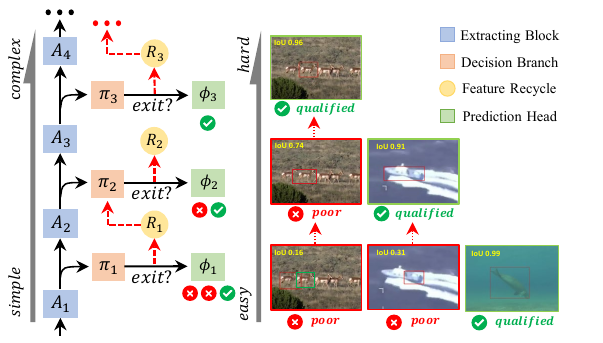}
	\vspace{-0mm}
	\caption{Efficient object tracking via instance-specific reasoning.}
	\vspace{-0mm}
	\label{fig:intro}
\end{figure}

In light of the above observations, we introduce the early exiting conception to tracking framework. 
Early exiting is proposed for instance-specific speed-up, it has been recently explored for vision~\cite{kaya2019shallow,cai2021dynamic} and language~\cite{pbee} tasks. 
Some methods~\cite{sun2021early,huang2017multi} attach additional classification heads, called internal classifiers, to select early layers of neural networks to shorten the inference time.
By this means, easy images (\eg cat and dog) can be learned by a compact model without redundant modules. In contrast, difficult images (\eg rare concepts, such as balloon) could be handled with more sophisticated network structures.
The demand of different model complexities for different instances indicates the potential of dynamic network structures, where the model speed can be dramatically improved while ensuring the accuracy.
Inspired by these works, 
in this paper, we propose DyTrack, a dynamic transformer tracking framework via instance-specific reasoning.
As shown in Figure~\ref{fig:intro}, 
we design decision branches (named decisioners) that 
are attached to intermediate layers of a transformer-stacked model. They control the termination of current forward propagation when the prediction is sufficiently reliable.
To improve the utilization efficiency of the model computation, we introduce a cascade feature recycling mechanism to reuse the features extracted from previous exits. 
Moreover, we propose a target-aware self-distillation strategy to jointly train the entire model, 
and features of the deep layer will serve as a teacher to supervise the early prediction, thus further improving the performance of early exits.
DyTrack is scalable for platforms with different computing resources.
Figure~\ref{fig:speed_vs_performance} shows the superiority of DyTrack which has favorable speed-precision trade-offs. 
The main contributions of this work are summarized as:

\begin{itemize}
	\vspace{-1mm}
	\item We propose DyTrack, the first dynamic transformer tracking framework.
	It offers a novel solution for efficient object tracking from the perspective of dynamic instance-specific computation. 
	
	\item A feature recycling mechanism is designed to reduce the redundancy and waste of computations. Moreover,
	we propose a target-aware self-distillation strategy which notably improves the precision of early predictions.
	
	\item Only training once, DyTrack can meet different speed-precision trade-offs 
	on different devices
	by simply changing the 
	dynamic routing conditions.
	DyTrack is scalable for platforms with different
	computing resources which gives it more application values.
	
	\item DyTrack is the first method that can meet both the high speed and high precision tracking demands using only a single model.
	Experiments demonstrate that it can achieve promising performance with good speed-precision trade-offs.
	For instance, 
	DyTrack runs at 256 $fps$ with 64.9\% AUC and 90 $fps$ with 69.2\% AUC on LaSOT.
\end{itemize}

\section{Related Work}

\noindent\textbf{Visual Object Tracking.} 
Siamese network~\cite{siamesefc} and discriminative correlation filter (DCF)~\cite{dcf} dominate the field of visual object tracking over the past decades. 
Tracking performance is constantly being refreshed with ongoing efforts.
SiamRPN~\cite{siameserpn} introduces the RPN~\cite{fast_rcnn} from detection into Siamese tracking framework, it gains accurate bounding boxes by offline training with large-scale image pairs
from ILSVRC~\cite{imagenet} and Youtube-BB~\cite{youtube-vos}. 
Then, SiamRNP++\cite{siamrpnplusplus} adopts deeper ResNet~\cite{resnet} for feature extraction, further enhancing the accuracy.
Online trackers~\cite{eco, ma2017robust, atom, dimp,zhang2023attention} obtain a clearer discriminative capability to interpret target and background 
by target-aware model fine-tuning.
For example, Zhang \etal~\cite{zhang2023attention} propose a attention-driven memory network for online tracking, in which an memory updater is designed to promote the model reliability. 
Recently, many trackers~\cite{transt,ostrack,mixformer,swintrack,grm} introduce transformer as their backbones~\cite{vit,swin}  and transformer-based template-search integration modules~\cite{attention_is_all}.
TransT~\cite{transt} proposes ego-context and cross-feature modules based on self-attention and cross-attention, respectively. 
More recently, many works~\cite{ostrack,mixformer,simtrack,vipt,swintrack,grm} take transformer backbone for joint feature extraction and interaction. 
MixFormer~\cite{mixformer} designs a backbone based on CVT~\cite{cvt} 
and OSTrack~\cite{ostrack} uses an MAE~\cite{mae} pre-trained ViT to joint extract the cues from the template and search region.
GRM~\cite{grm} designs a token division module to select search tokens to interact with the template tokens for alleviating target-background confusion, and more complex modeling relationships are learned.
From the perspective of training data, Li \etal~\cite{li2023self} propose to synthesize sufficient samples by a crop-transform-paste operation,  boosting the performance of various supervised trackers.
However, most works focus on the pursuit of higher 
performance while ignoring efficiency which should also be critical in practical applications.

\noindent\textbf{Efficient Tracking.} 
In real-world applications, tracking algorithms usually require high efficiency for diverse deployment purposes, such as unmanned vehicles and UAV vision.
Some studies have explored approaches to achieve efficient tracking.
ECO~\cite{eco} introduces a factorized convolution operator to reduce the computational complexity in DCF models. 
It has a high inference speed, but the performance 
is inferior to meet the current tracking requirement.
To meet the requirements of resource-limited platforms for model FLOPs and parameters, LightTrack~\cite{lighttrack} uses NAS~\cite{pham2018efficient} 
to search for a tracker with a lightweight backbone and prediction head.
While in practice, FLOPs and the parameter number do not directly measure the actual speed of the model ~\cite{shufflenetv2}.
Ussa \etal~\cite{ussa2023hybrid} propose a real-time hybrid neuromorphic tracker that aggregates clues from event cameras for maintaining robust tracking in occluded scenarios.
More recently, FEAR~\cite{fear} employs depth-wise separable convolutions and 
designs compact feature extraction and fusion blocks,
and E.T.Track~\cite{ettrack} proposes a single instance-level attention layer to build a transformer-based tracker. They both attain high energy efficiency.
However, like high-performance trackers but at another extreme, these methods favor efficiency in the efficiency-precision trade-off. 
They remain constrained in dealing with some complex and corner cases.
LiteTrack~\cite{litetrack} employs asynchronous feature extraction and interaction, and reduces redundant computation through the proposed encoder layer pruning.  
MixFormerV2~\cite{mixformerv2} proposes a fully transformer-based tracking framework that eliminates dense convolutional operations and complex score heads, achieving real-time speed on CPU platform. 
To use large-stride downsampling hierarchical backbone for tracking, HiT~\cite{hit} incorporates the high-level information of deep features into the shallow large-resolution features. 
ABTrack~\cite{abtrack} designs adaptively bypassing transformer and introduce a pruning method to speed up tracker inference.
For efficient multi-modal tracking, Zhang et al.~\cite{zhang2023efficient} introduce a cross-modality distillation framework that transfers effective knowledge from a powerful two-stream tracker to a compact single-stream tracker.
These approaches improve efficiency by designing lighter architectures or utilizing pruning and distillation techniques. However, they are fixed-speed trackers tailored to specific platforms. When adapting to different speed-precision trade-offs, they require re-training or even re-designing the model.
Based on SiamFC~\cite{siamesefc}, EAST~\cite{east} achieves adaptive tracking with 
cascaded CNN features.
Despite achieving a promising speedup, it needs to learn policies by complex reinforcement learning and has to train different models for different precision-speed trade-offs which lacks flexibility and utility.
In contrast, we try to address the above concerns from the perspective that we formulate the speed-precision problem in a dynamic transformer network by instance-specific routing.

\noindent\textbf{Instance-specific Computation.}
Instance-specific computation methods~\cite{skipnet,veit2018convolutional,blockdrop,dvt} learn to dynamically assign computing resources on a per-input basis.
Wang~\etal~\cite{skipnet} observe that shallow networks are sufficient for many inputs, and they develop SkipNet which uses a gating network to selectively skip convolutional blocks to reduce computation.
SDN~\cite{kaya2019shallow} adds internal classifiers to select early layers to exit, shortening the processing time. 
PBEE~\cite{pbee} proposes a patience-based early exiting method that exits for early layers when obtaining $t$ consecutive unchanged answers, showing higher efficiency on NLP tasks.
ZTW~\cite{ztw} introduces cascade connections and aggregates predictions from all previous predecessors by an extra weighted geometric mean. 
Some methods~\cite{gaternet,abati2020conditional} implement more customized gating operations to control the number of channels assigned to each input instance.
Recently, DVT~\cite{dvt} proposes to adaptively assign a decent token number conditioned on each image to gain higher computational efficiency.
Early exiting approaches are also applied to some downstream tasks, such as image captioning~\cite{deecap} and video recognition~\cite{frameexit}.
Drawing inspiration from these works, we focus on the family of early exiting and address to bring its advantages in efficient reasoning to visual object tracking.

\begin{figure*}[!t]
	\centering
	\includegraphics[width=0.9375\textwidth]{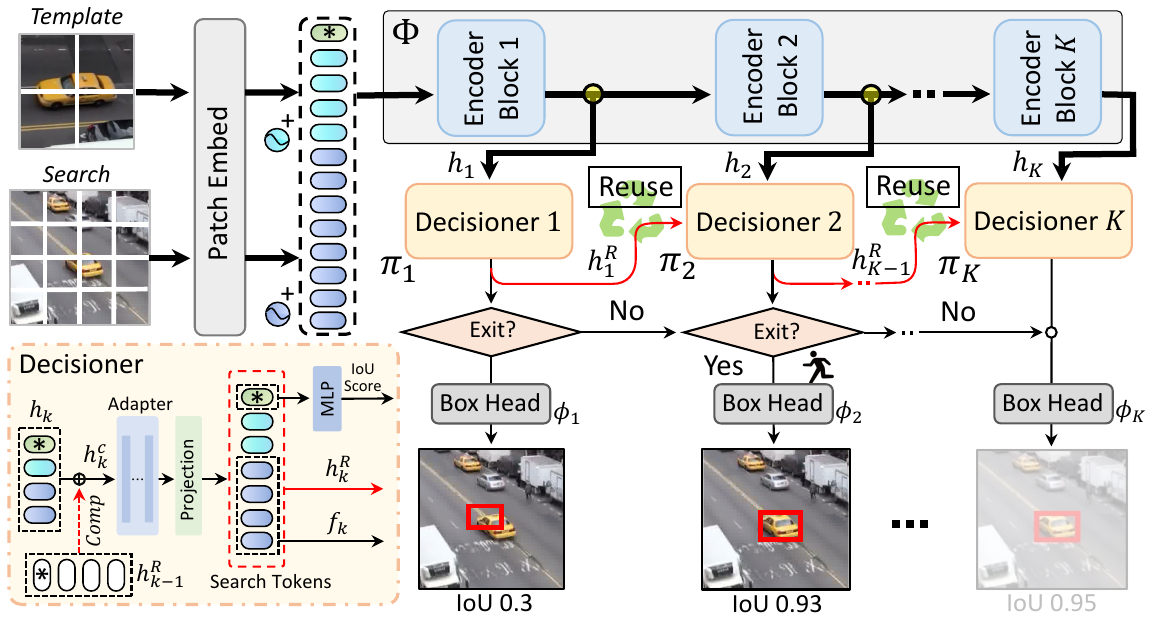}
	\vspace{-0mm}
	\caption{An overview of DyTrack. This framework achieves instance-specific inference where the forward propagation will terminate early when the current feature representation is reliably conditioned on the input. We take the learned IoU score as the condition to determine whether the prediction is sufficiently confident, and the cascade feature recycling mechanism allows reusing computation of the predecessors.
		DyTrack can achieve promising speed-precision trade-offs.}
	\vspace{-0mm}
	\label{fig:overview}
\end{figure*}


\section{Dynamic Transformer Tracking}
\label{sec:method}

A typical visual object tracker consists of feature extraction, template-search region interaction, and a box prediction head. 
Deeper backbones (\eg ResNet~\cite{resnet} and ViT~\cite{vit}) and more complex interaction mechanisms (\eg cross attention~\cite{attention_is_all}) are widely employed to accomplish the tracking modeling.
Both the high-performance and high-efficiency trackers achieve satisfactory results which they aim to achieve,
but they are designed with static inference routes that treat all inputs equally.
According to our observation (Figure~\ref{fig:motivation}), the complexity of tracking scenarios is diverse, with only a limited proportion of difficult ones.
To this end, in this work, we propose DyTrack to execute instance-specific computations.
DyTrack assigns simple inference routes to easy frames, saving computational resources 
along with accelerating the inference.
When facing difficult cases, it can assign complex networks to obtain accurate tracking.

\subsection{Overview}
Given the initial position of the target object in a video sequence, a tracker aims to predict the bounding box in all subsequent frames. 
For our DyTrack, it simultaneously needs to terminate forward inference when its confidence score $\{s_k\}_{k=1}^{K}$ is found to be reliable enough.
The overall framework is shown in Figure~\ref{fig:overview}, which consists of
$\textit{\textbf{\romannumeral1})}$ a feature extraction and interaction network $\Phi$; 
$\textit{\textbf{\romannumeral2})}$ $K$ termination operation decisioners $\{\pi_k\}^{K}_{k=1}$;
$\textit{\textbf{\romannumeral3})}$ $K$ corresponding box prediction heads $\{\phi_{k}\}^{K}_{k=1}$.
For the input template  $\boldsymbol{Z} \in \mathbb{R}^{3\times H_z \times W_z} $ and search region $\boldsymbol{X} \in \mathbb{R}^{3\times H_x \times W_x} $, we use $\Phi$ to obtain the independent representation of the current frame.
When the features propagate to specified nodes, the decisioners $\pi_k$ attached to the hidden layers will discriminate them, and if they are predicted to be reliable enough, they will be directly input to the current box head to return a box result, and the inference terminates.
Otherwise, the features will continue to propagate to the next exiting node to adopt deeper routes until receiving a halt signal. 
An inference process is represented as:
\begin{align}
	\label{eq:overview}
	\bm{B} &= \phi_k(\pi_k(\Phi(\boldsymbol{Z}, \boldsymbol{X}))), {k =  \underset{k}{\text{min}}\{k~|~s_k>\tau_k\}},
\end{align} 
where $\bm{B}$ is the predicted bounding box, and $k$ denotes the earliest node that meets the exiting condition $\tau_k$.

\noindent\textbf{Feature Extraction and Box Head.}
Pure transformer~\cite{mixformer,ostrack,swin,simtrack} architecture recently replaces the CNN + transformer fusion~\cite{transt, stark} in object tracking.
These methods are implemented with a one-stream architecture which allows the template and search region to be dependent on each other from the first network layer.
One-stream paradigm also simplifies the tracking framework which facilitates the introduction of dynamic reasoning.
Typical siamese trackers~\cite{siamesefc, siameserpn} adopt separate feature extracting and interaction, a valid round of inference must pass through these two components successively, which is not conducive to the model exiting in the early stage. 
Therefore, we follow the one-stream paradigm and employ a ViT~\cite{vit} backbone for feature extraction and interaction.
The input template $\boldsymbol{Z}$ and search region $\boldsymbol{X}$ are first split into patches, projected and flattened to token sequences $\boldsymbol{O}_z^0 \in \mathbb{R}^{N_{z}\times D}$ and $\boldsymbol{O}_x^0 \in \mathbb{R}^{N_{x}\times D}$. 
Accordingly, they are added to the position embeddings.
Hereby, inspired by the class token in ViT, we introduce an IoU token  $\boldsymbol{O}_s^0 \in \mathbb{R}^{1\times D}$ for IoU score prediction.
The concatenated token sequences $\boldsymbol{O}^0 = [\boldsymbol{O}_s^0, \boldsymbol{O}_z^0, \boldsymbol{O}_x^0]$ are fed into stacked transformer layers.
After propagating in the backbone encoder and the decisioner networks, we feed the search tokens into a box prediction head~\cite{stark} $\phi_k$ to return the results. Specially, we replace the 3x3 convolutional layers with RepVGG~\cite{repvgg} blocks to improve the inference efficiency.
Here, the transformer backbone and box head compose our baseline model.

\noindent\textbf{Decisioner Network.}
In image classification, the decisioner module can be easily served by several fully connected layers because the top-1 score is the only evaluation criterion. While tracking task needs to predict the bounding-box of the target object, 
more suitable module needs to be designed.
In DyTrack, we attach 
the designed decisioner networks $\{\pi_{k}\}_{k=1}^{K}$ to $K$ hidden layers 
of the backbone to evaluate whether the feature is reliable.
As shown in Figure~\ref{fig:overview}, the input features $\bm{h}_k$ and the reused features $\bm h_{k-1}^{R}$ are synthesized to composite features:
\begin{align}
	\label{eq:reuse_summary}
	\bm{h}_{k}^{c} &= Comp(\bm h_k, \bm h_{k-1}^{R}),
\end{align} 
where $Comp$ represents the feature reuse manner (detailed descriptions in Section~\ref{sec:FR}). 
We keep the shape and dimension of $\bm{h}_{k}^{c}$ the same as $\bm h_k$.
Note that $\pi_1$ only receives $\bm h_1$ as the input.
Then the features are fed into an adapter block to learn a paradigm shift for discriminating the target object. 
The adapter is composed of several transformer layers. After a linear projection operation, the generated search tokens serve as candidates for the inputs to the box prediction head.
Following previous studies~\cite{transt-m, artrackv2}, the IoU token, is fed into a 3-layer MLP to estimate the current frame's IoU score which is the exiting criterion.

\subsection{Feature Recycling}
\label{sec:FR}

\begin{figure}[t]
	\centering
	\includegraphics[width=0.46\textwidth]{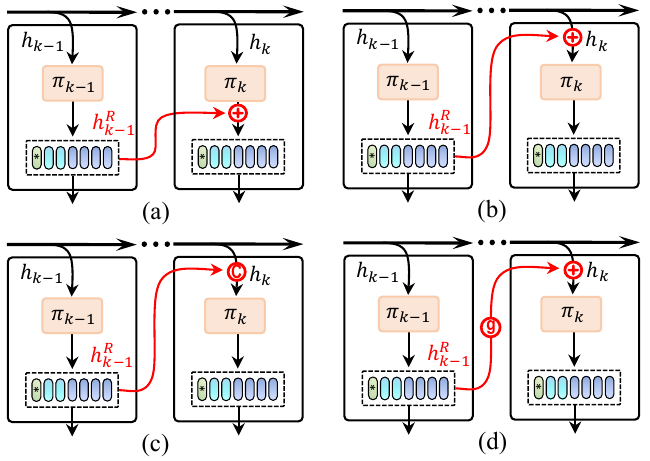}
	\vspace{-0mm}
	\caption{Diagrams of four feature reuse schemas. The output of the score head is omitted for better readability.}
	\vspace{-2mm}
	\label{fig:feat_reuse}
\end{figure}

In an effort to reduce the computation waste and redundancy, we introduce a feature recycling mechanism to reuse the features generated by the predecessors.
Concretely, we study four variants of reuse manner as shown in Figure~\ref{fig:feat_reuse}.
(\textbf{a}) The recycled features $\bm h_{k-1}^{R}$ are integrated into the current outputs of the decisioner network $\pi_k$ through element-wise summation. This manner, in other words, alleviates the pressure of $\pi_k$ and it only needs to learn the residual with respect to $\bm h_{k-1}^{R}$.
(\textbf{b}) The recycled features $\bm h_{k-1}^{R}$ are integrated into the current inputs of 
$\pi_k$ through element-wise summation to enhance the input features.
(\textbf{c}) 
The recycled features 
$h_{k-1}^{R}$ are integrated into the current inputs of 
$\pi_k$ through token-wise (the first dimension) concatenation.
(\textbf{d}) Similar to (\textbf{b}), but a residual gate~\cite{yang2022lavt} is attached to $\bm h_{k-1}^{R}$ to select valuable information.  
In practice, the variants (\textbf{c}) and (\text{d}) introduce extra computations. 
And we find (\textbf{b}) is superior to (\textbf{a}) with the same computational burden.
Therefore, in our work, we adopt the feature reuse schema (\textbf{b}) as the default manner in DyTrack.

\subsection{Target-aware Self-distillation}
\label{sec:tasd}
\begin{figure}[!t]
	\centering
	\includegraphics[width=0.4825\textwidth]{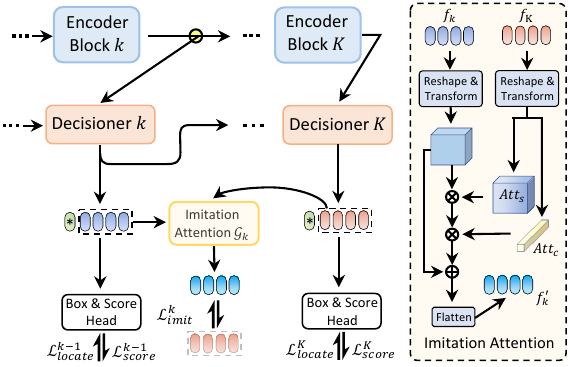}
	\vspace{-1mm}
	\caption{Illustration of the target-aware self-distillation.
		Early features are learned with additional supervision from teacher from the deep layer. Imitation attention re-weight the student features to valuable regions in terms of teacher features.
		We omit the output of the score head for better readability.}
	\vspace{-2mm}
	\label{fig:self_dist}
\end{figure}

DyTrack has multiple exiting routes to execute box prediction. 
For deeper layers, it is usually easier to learn high-level semantic representation~\cite{deecap}.
Hence, the earlier the output node is, the harder an accurate tracking branch can be learned.
To bridge this gap, we propose to encourage the shallow and middle branches to mimic the response pattern of the deep layer.
Specifically, we introduce a target-aware self-distillation training strategy whose process is shown in Figure~\ref{fig:self_dist}.
In DyTrack,
the deepest feature representation $\bm{f}_K$ acts as the teacher of all previous features $\{\bm{f}_k\}_{k=1}^{K-1}$,
and multiple prediction branches distributed in various depths of the network enable conditions for self-distillation learning.
It may not be optimal to equally distill all the features. 
Intuitively, the object response regions are more critical than the background regions.
To further promote the distillation, we introduce an imitation attention module $\mathcal{G}_k$ to re-weight the learned features of different regions, where $k \in [1,K-1]$.
The imitation attention mechanism is achieved by re-weighting the student features, which is implemented through spatial and channel attention derived from the teacher features. 
In this way, the student features can focus more effectively on the key regions and channels during the distillation process.
Specifically, $\mathcal{G}_k$ receives two inputs which are student tokens $\bm f_k$ from hidden branches and teacher tokens $\bm f_K$ from finally reasoning route.
After respective reshape and transform (convolutional and linear layers) operations, $\bm f_K$ is projected to two groups of spatial attention maps $\bm{Att}_s$ and channel attention vectors $\bm{Att}_c$. 
Specifically, the transform operation including a 1$\times$1 convolution layer and a linear projection layer. To obtain $\bm{Att}_c$, we additionally employ a $softmax$ operation in the spatial dimension of the features. 
These learned attention acts on the student features to obtain the target-aware feature representation $\bm f_k^{'}$.
Finally, we utilize the cosine similarity as the imitation measurement between the teacher representation $\bm f_{K}$ and the re-weighted student representation $\bm f_k^{'}$ as: 
\begin{align}
	\label{eq:cos_sim}
	Cos\text{-}sim(\bm f_k^{'}, \bm f_{K}) &= 1-\frac{\bm f_k^{'} \cdot \bm f_{K} }{||\bm f_k^{'}||\cdot|| \bm f_{K}|| },
\end{align} 
where $||\cdot||$ denotes the $\mathcal{L}_2$ norm.
We enumerate all hidden branches, resulting in a feature imitation loss as:
\begin{align}
	\label{eq:mimic_loss}
	\mathcal{L}_{imit} &= \frac{1}{K-1}\sum_{k=1}^{K-1}Cos\text{-}sim(\bm f_k^{'}, \bm f_{K}),
\end{align} 
Without extra teachers, the target-aware self-distillation process are trained end-to-end. 
And note that, at the inference, the imitation attention modules will be removed without bringing additional computations. 

\subsection{Training Objective}
For object locating, we combine the $\mathcal{L}_1$ loss and the generalized GIoU loss~\cite{giou} $\mathcal{L}_G$ as the locating loss, which can be formulated as:
\begin{align}
	\label{eq:locate_loss}
	\mathcal{L}_{locate} &= \lambda_1\mathcal{L}_1(\bm B, \bm{B}_{gt}) + \lambda_G\mathcal{L}_G(\bm B, \bm{B}_{gt}),
\end{align} 
where $\bm{B}_{gt}$ represents the ground truth, $\lambda_1=5$ and $\lambda_G=2$ are the weight parameters 
as setting in ~\cite{stark}.
For IoU score prediction, we adopt a conventional $\mathcal{L}_2$ loss.
The overall 
joint 
training objective is formulated as:
\begin{align}
	\label{eq:locate_loss}
	\mathcal{L}_{joint} &= \mathcal{L}_{locate} + \lambda_{s}\mathcal{L}_{score} + \lambda_m\mathcal{L}_{imit}.
\end{align} 
where $\lambda_{s}=5$ and $\lambda_m=10$ in our experiments.
DyTrack is a “once-and-for-all” model which only need to train once and can obtain various speed-precision trade-offs by simply adjusting the termination conditions.

\subsection{Pareto Optimality}
Speed-precision trade-off is a typical optimizing problem of multiple objectives. 
Compromises have to be made since no single solution can optimize each objective simultaneously.
In this work, we use Pareto optimization over speed and precision to evaluate trackers.
A solution is said to be Pareto-optimality if none of the objectives can be improved without worsening any other objective, and the Pareto-optimality solutions are said to be in the Pareto front.

\section{Experiments}

\begin{table*}[!t]
	\caption{State-of-the-art comparison on large-scale 
		LaSOT~\protect\cite{lasot},
		LaSOT$_{ext}$~\protect\cite{lasotext}, 
		GOT-10k~\protect\cite{got10k}, 
		and TrackingNet~\protect\cite{trackingnet} 
		test benchmarks.
		The symbol * indicates training on the train split of GOT-10k.
		The best two results are shown in \textcolor{red}{{red}} and \textcolor{blue}{{blue}} fonts. The Pareto-optimality items in terms of performance and speed are shown in \textbf{\textit{italic}} font.}
	\label{tab:compare_sota}
	\vspace{-0mm}
	\centering
	\renewcommand\arraystretch{1.2}
	\small 
	\resizebox{0.98\linewidth}{!}{
		\setlength{\tabcolsep}{1.5mm}{
			\begin{tabular}{cc|ccc|ccc|ccc|ccc|c}
				\hline	
				\multicolumn{2}{c}{\multirow{2}{*}{Method}} &  \multicolumn{3}{c}{LaSOT} & \multicolumn{3}{c}{LaSOT$_{ext}$} & \multicolumn{3}{c}{GOT-10k*} & \multicolumn{3}{c}{TrackingNet} &\multirow{2}{*}{\tabincell{c}{Speed\\(fps) }}\\
				\cline{3-14}
				
				& & AUC & P$_{Norm}$ & P & AUC & P$_{Norm}$ & P & AO & SR$_{50}$ & SR$_{75}$ & AUC & P$_{Norm}$ & P & \\
				\hline  
				\multicolumn{2}{l}{SiamFC~\cite{siamesefc} } \vline
				&33.6&42.0&33.9 &23.0&31.1&26.9  &34.8 &35.3 &9.8  &57.1&66.3 &53.3 &100 \\
				\multicolumn{2}{l}{ECO~\cite{eco} }\vline
				&32.4 &33.8 &30.1 &22.0&25.2&24.0 &31.6 &30.9 &11.1 &55.4 &61.8 &49.2&{\textcolor{blue}{{240}}}  \\
				\multicolumn{2}{l}{SiamRPN++~\cite{siamrpnplusplus}}\vline
				&49.6 &56.9 &49.1   &34.0&41.6&39.6    &51.7 &61.6 &32.5 &73.3 &80.0 &69.4&56\\
				\multicolumn{2}{l}{ATOM~\cite{atom}}\vline 
				&51.5 &57.6 &50.5  &37.6&45.9&43.0   &55.6 &63.4 &40.2 &70.3 &77.1 &64.8&83\\
				\multicolumn{2}{l}{DiMP~\cite{dimp}}\vline
				&56.9 &65.0 &56.7  &39.2&47.6&45.1   &61.1 &71.7 &49.2 &74.0 &80.1 &68.7&77\\
				\multicolumn{2}{l}{OCEAN~\cite{ocean}}\vline 
				&56.0 &65.1 &56.6 &- &- &- &61.1 &72.1 &47.3 &-&-&-&25 \\
				\multicolumn{2}{l}{PrDiMP~\cite{prdimp}}\vline
				&59.8 &68.8 &60.8  &- &-&- &63.4 &73.8 &54.3 &75.8 &81.6 &70.4&47\\
				\multicolumn{2}{l}{SiamR-CNN~\cite{siamrcnn}}\vline
				&64.8&72.2&-  &-&-&-  &64.9 &72.8 &59.7 &81.2 &85.4 &80.0&5\\ 
				\multicolumn{2}{l}{TrSiam~\cite{trdimp}}\vline
				&62.4 &- &60.6   &- &-&-  &67.3 &78.7 &58.6 &78.1 &82.9 &72.7&40\\
				\multicolumn{2}{l}{LightTrack~\cite{lighttrack}}\vline
				&53.8 &- &53.7  &-&-&-   &61.1 &71.0 &- &72.5 &77.8 &69.5&128\\
				\multicolumn{2}{l}{TrDiMP~\cite{trdimp}}\vline 
				&63.9 &- &61.4  &-&-&- &67.1 &77.7 &58.3 &78.4 &83.3 &73.1&41\\
				\multicolumn{2}{l}{TransT~\cite{transt}}\vline
				&{64.9} &{73.8} &{69.0} &45.1&51.3&{{51.2}} &67.1 &76.8 &60.9 &{81.4} &{86.7} &80.3&63\\
				\multicolumn{2}{l}{STARK-ST50~\cite{stark}}\vline
				&66.4&-&71.2 &-&-&-   &68.0&77.7&62.3&81.3&86.1&-&50\\
				\multicolumn{2}{l}{FEAR~\cite{fear}}\vline
				&53.5 &- &54.5 &- &- &- &61.9 &72.2 &- &-&-&-&105 \\
				\multicolumn{2}{l}{CSWinTT~\cite{cswintt}}\vline 
				&66.2 &75.2 &70.9 &-&-&-  &{69.4} &78.9 &{65.4}  &81.9 &{86.7} &79.5&12\\ 
				\multicolumn{2}{l}{SBT~\cite{sbt}}\vline
				&65.9 &- &70.0 &-&-&-&69.9&\textcolor{blue}{80.4}&63.6&-&-&-&37\\
				\multicolumn{2}{l}{AiATrack~\cite{aiatrack}}\vline
				&69.0 &\textbf{\textit{\textcolor{red}{79.4}}} &73.8 &\textcolor{blue}{47.7}&\textcolor{blue}{55.6}&\textbf{\textit{\textcolor{red}{55.4}}} &69.6&{80.0}&63.2&82.7&\textcolor{blue}{87.8}&80.4&38\\
				\multicolumn{2}{l}{MixFormer~\cite{mixformer}}\vline 
				&\textcolor{red}{{69.2}} &{78.7} &\textcolor{blue}{{74.7}} &-&-&- &{{70.7}}&{{80.0}}&{{67.8}}&\textbf{\textit{\textcolor{red}{{83.1}}}}&\textbf{\textit{\textcolor{red}{{88.1}}}}&{\textit{\textcolor{red}{\textbf{81.6}}}}&25\\
				\multicolumn{2}{l}{E.T.Track~\cite{ettrack}}\vline
				&59.1 &- &-  &-&-&-  &- &- &- &70.5 &80.3 &70.6&40 \\
				\multicolumn{2}{l}{HCAT~\cite{hcat}}\vline
				&59.3 &68.7 &61.0   &-&-&-   &\textbf{\textit{65.1}} &76.5 &56.7 &76.6 &82.6 &72.9&195 \\
				\multicolumn{2}{l}{MAT~\cite{mat}}\vline
				&\textbf{\textit{{67.8}}} &\textbf{\textit{{77.3}}} &{-} &-&-&- &\textbf{\textit{67.7}} &\textbf{\textit{78.4}} &- &\textbf{\textit{81.9}} &\textbf{\textit{86.8}} &-&111\\
				\multicolumn{2}{l}{CTTrack-B~\cite{cttrack}}\vline
				&67.8 &{77.8} &{-} &-&-&- &\textcolor{blue}{71.3} &\textbf{\textit{\textcolor{red}{80.7}}} &\textit{\textcolor{red}{\textbf{70.3}}} &82.5 &87.1 &80.3 &40\\
				\hline
				\multirow{3}{*}{{\ \ \ \ \ DyTrack \ \ \ }}
				&Fast &\textbf{\textit{64.9}}&\textbf{\textit{73.4}}&\textbf{\textit{67.8}} &\textbf{\textit{44.0}}&\textbf{\textit{52.2}}&\textbf{\textit{47.6}}     &\textbf{\textit{65.0}}&\textbf{\textit{73.4}}&\textbf{\textit{57.9}}&\textbf{\textit{79.4}}&\textbf{\textit{83.7}}&\textbf{\textit{75.8}}&\textit{\textcolor{red}{\textbf{256}}}\\
				\cline{2-15}
				&Medi &\textbf{\textit{66.5}}&\textbf{\textit{75.5}}&\textbf{\textit{70.4}}&\textbf{\textit{{{46.3}}}}&\textbf{\textit{{{55.4}}}}&\textbf{\textit{51.1}}&\textbf{\textit{66.2}}&\textbf{\textit{75.2}}&\textbf{\textit{59.4}}&\textbf{\textit{80.9}}&\textbf{\textit{85.5}}&\textbf{\textit{77.8}}&\textit{{\textbf{196}}}\\
				\cline{2-15}
				&Base&\textit{\textcolor{red}{\textbf{69.2}}} &\textbf{\textit{\textcolor{blue}{{78.9}}}} &\textbf{\textit{\textcolor{red}{{75.2}}}}  &\textbf{\textit{\textcolor{red}{{48.1}}}} &\textbf{\textit{\textcolor{red}{{58.1}}}} &\textbf{\textit{\textcolor{blue}{{54.6}}}}  &\textbf{\textit{\textcolor{red}{{71.4}}}} &\textbf{\textit{{{80.2}}}} &\textbf{\textit{\textcolor{blue}{{68.5}}}}  &\textbf{\textit{\textcolor{blue}{{82.9}}}} &\textbf{\textit{{{87.3}}}} &\textbf{\textit{{\textcolor{blue}{81.2}}}}&\textit{{\textbf{90}}}\\
				
				\hline
	\end{tabular}}}
	\vspace{-2mm}
\end{table*}

\subsection{Implementation Details}
\label{sec:implem_detail}

\noindent\textbf{Model.}
We report DyTrack with three speed-precision trade-offs, as shown in Table~\ref{tab:compare_sota}.
The Fast, Medi, and Base versions can run at 256, 196, and 90 $fps$, respectively.
We adopt ViT-B~\cite{vit} with MAE~\cite{mae} pre-trained parameters as our encoder network.
We set the number of exiting nodes $K$ to 3, which we find more efficient in practice.
These nodes exist in the output layers of the 2nd, 6th, and 12th of the encoder. 
The adapters of decisioner networks consist of 2, 1, and 0 transformer layers in the the 1st, 2nd, and 3rd exiting branches, respectively.
The speed of the trackers is tested on 
a single Nvidia 2080Ti GPU
with a batch size of 1 (we also provide results on edge computing platform in the following subsection).

\noindent\textbf{Training Details.}
We employ the training splits of GOT-10k~\cite{got10k}, TrackingNet~\cite{trackingnet}, COCO~\cite{coco}, and LaSOT~\cite{lasot} as our training data. 
For the evaluation of GOT-10k, we only use the training split of GOT-10k to follow the official requirements.
We sample from the training sequence to generate the image pairs. 
For COCO which is an imageg dataset, we execute different data augmentations on the same image to obtain the image pairs.
The search region and template size are $4^2$ and $2^2$ the area of the object, which are resized to 256$\times$256 and 128$\times$128 patches before fed into the model.
We train the model in two stages: we first train the backbone and the final prediction branch, then we include the other exiting branches for joint training. 
The training configurations of the two stages are the same.
AdamW~\cite{adamw} is employed as our optimizer, with a weight decay of 1e-4. 
The learning rate of the backbone is set to 1e-5, 
and the remaining 
modules are set to 1e-4.
We use 2 Nvidia A100 GPUs to train our models.
In each stage, the model is trained for 300 epochs with a batch size of 64.
Each epoch contains 60,000 sampling pairs.
The learning rate decreases by a factor of 10 after 240 epochs.

\subsection{State-of-the-art Comparisons}
We compare DyTrack with an extensive list of state-of-the-art methods which include real-time and non-real-time trackers on seven widely used tracking benchmarks.

\noindent\textbf{LaSOT.}
LaSOT~\cite{lasot} is a long-term dataset including 280 test sequences of 14 challenge categories.
As reported in Table~\ref{tab:compare_sota}, DyTrack-Fast achieves 256 fps with an AUC of 64.9\%, which is as good as TransT~\cite{transt}. 
While TransT only runs at 63 fps (4.1$\times$ slower). 
DyTrack-Medi obtains 66.5\% AUC with a high speed of 196 fps.
With a similar speed, HCAT~\cite{hcat} 
only attends 59.3\% AUC.
With a similar performance, STARK-ST50~\cite{stark} 
only attends 50 fps. 
Furthermore, DyTrack-Base ranks first with an AUC of 69.2\%, with the same AUC score as MixFormer, while performing much faster (90 $vs$ 25 $fps$). 
All our models achieve Pareto-optimality with superior speed-precision trade-offs.

\begin{table}[!t]
	\centering
	\renewcommand\arraystretch{1.1}
	\small 
		\resizebox{0.95\linewidth}{!}{
			\setlength{\tabcolsep}{2mm}{
				\begin{tabular}{cc|ccc|c}
					\hline	
					\multicolumn{2}{c}{Method} \vline &TNL2k&NFS&UAV123&Speed \\
					\cline{3-6}
					
					\hline  
					\multicolumn{2}{c}{SiamFC~\cite{siamesefc}}\vline
					&29.5 &37.7 &46.8 &100 \\
					\multicolumn{2}{c}{ECO~\cite{eco} }\vline
					&32.6 &52.2 &53.5 &{\textcolor{blue}{{240}}} \\
					\multicolumn{2}{c}{ATOM~\cite{atom}}\vline 
					&40.1 &58.3 &63.2&83 \\
					\multicolumn{2}{c}{DiMP~\cite{dimp}}\vline 
					&44.7& 61.8&64.3&77 \\
					\multicolumn{2}{c}{SiamRPN++~\cite{siamrpnplusplus}}\vline
					&41.3 &50.2 &61.6 &56\\
					\multicolumn{2}{c}{OCEAN~\cite{ocean}}\vline
					&38.4 &49.4 &57.4&25 \\
					\multicolumn{2}{c}{LightTrack~\cite{lighttrack}}\vline
					&- &55.3 &62.5 &128 \\
					\multicolumn{2}{c}{TransT~\cite{transt}}\vline
					&50.7 &65.7 &\textcolor{blue}{69.1}&63\\
					\multicolumn{2}{c}{STARK-ST50~\cite{stark}}\vline
					&-&64.3&68.4&50 \\
					\multicolumn{2}{c}{FEAR~\cite{fear}}\vline
					&- &61.4 &-&105 \\
					\multicolumn{2}{c}{Sim-B/16~\cite{simtrack}}\vline 
					&\textcolor{blue}{54.8} &- &\textbf{\textit{\textcolor{red}{69.8}}} &87 \\
					\multicolumn{2}{c}{HCAT~\cite{hcat}}\vline
					&- &63.5 &62.7 &195 \\
					\multicolumn{2}{c}{E.T.Track~\cite{ettrack}}\vline
					&- &59.0 &62.3 &40 \\
					\multicolumn{2}{c}{MAT~\cite{mat}}\vline
					&51.3 &65.3 &68.0 &111 \\
					\hline
					\multirow{3}{*}{{\ \ \ \ \ DyTrack \ \ \ }} &Fast &\textbf{\textit{50.7}}&\textbf{\textit{63.3}} &\textbf{\textit{66.7}}&\textit{\textcolor{red}{\textbf{256}}}\\
					\cline{2-6}
					&Medi  &\textbf{\textit{53.8}}&\textbf{\textit{\textcolor{blue}{66.5}}} &\textbf{\textit{68.2}}&\textit{{\textbf{196}}}\\
					\cline{2-6}
					& Base&\textbf{\textit{\textcolor{red}{{55.6}}}} &\textbf{\textit{\textcolor{red}{{66.7}}}} &\textbf{\textit{\textcolor{blue}{{69.1}}}}&\textit{{\textbf{90}}}\\
					\hline
		\end{tabular}}}
	\vspace{-0mm}
	\caption{SOTA comparison on TNL2k~\cite{tnl2k}, NFS~\cite{nfs}, and UAV123~\cite{uav} test benchmarks in AUC score and $fps$.}
	\label{tab:compare_sota2}
	\vspace{-1mm}
\end{table}

\noindent\textbf{LaSOT\_{ext}.}
LaSOT\_{ext}~\cite{lasotext} is an extension dataset of LaSOT~\cite{lasot}, which includes more challenge videos.
We test our models on its 150 videos sequences.
The results are reported in Table~\ref{tab:compare_sota}.
DyTrack-Base gets the best AUC of 48.1\%, surpassing the previous best tracker AiATrack~\cite{aiatrack}.
DyTrack-Medi also obtains an AUC of 46.3\% with a speed of 196 $fps$, which performs better and faster than the recent published transformer tracker TransT
(45.1\% AUC with 63 $fps$).

\noindent\textbf{GOT-10k.}
GOT-10k~\cite{got10k} is a large-scale dataset with rich object categories and movement modes.
The object categories of train split and test split are zero-overlapped.
Following the official requirements, the models are only trained with GOT-10k training split.
The results are reported in Table~\ref{tab:compare_sota}, Dytrack-Base obtains the best AO score of 71.4\%, outperforming the previous state-of-the-art method CTTrack~\cite{cttrack}. 
DyTrack also arrives the Pareto-optimality.

\noindent\textbf{TrackingNet.}
TrackingNet~\cite{trackingnet} is a large-scale dataset containing 30,643 sequences (30,132 for training and 511 for testing). 
The evaluation is provided by the official online 
server following the one-pass evaluation protocol.
As reported in Table~\ref{tab:compare_sota}, DyTrack-Base 
obtains a competitive performance
with a speed of 90 $fps$, while the SOTA AUC is 
0.2\% better than ours. DyTrack also makes the Pareto-optimality with promising speed-precision trade-offs, \eg DyTrack-Medi gets an AUC score of 80.9\% with a speed of 196 $fps$, which is much faster than other algorithms with the same accuracy.

\noindent\textbf{TNL2k.}
TNL2k~\cite{tnl2k} is a recently released dataset containing 700 test videos. 
The results are presented in Table~\ref{tab:compare_sota2}.
DyTrack-Base performs the best among all compared trackers with an AUC score of 55.6\%. Meanwhile, it is also slightly ahead of the second best tracker Sim-B in terms of speed.
DyTrack-Medi and DyTrack-Fast also achieve impressive performance, with a substantial speed lead.

\noindent\textbf{NFS and UAV123.}
NFS~\cite{nfs} consists of 100 challenging videos captured using a high frame rate camera. We evaluate our model on the 30 $fps$ version of this dataset.
UAV123~\cite{uav} provides 123 aerial videos captured from a UAV platform. 
These two datasets are small-scale benchmarks.
As shown in Table~\ref{tab:compare_sota2},
our DyTrack-Base obtains the best AUC score on NFS as reported in Table~\ref{tab:compare_sota2}.
On UAV123, DyTrack-Base also ranks second with an AUC score of 69.1\%.
DyTrack-Medi and DyTrack-Fast also get satisfying results with higher speeds, which achieve Pareto-optimality.

\begin{table}[b]
	\renewcommand\arraystretch{1.3}
	\definecolor{purple(x11)}{rgb}{0.63, 0.36, 0.94}
	\definecolor{yellow(munsell)}{rgb}{1.0,0.988, 0.957}
	\definecolor{green(colorwheel)(x11green)}{rgb}{0.0, 1.0, 0.0}
	\definecolor{pink}{rgb}{1.0, 0.85, 0.85}
	\vspace{-0mm}
	\centering
	\small
	\resizebox{0.98\linewidth}{!}{
		\setlength{\tabcolsep}{2.75mm}{
			\begin{tabular}{l|c|ccc|c}
				\hline
				\# & Method &E-1 &E-2 &E-3 &$\Delta$\\
				\hline
				1 & DyTrack &64.4 &67.1&69.2 &--\\
				\hdashline
				2 &$w/o$ Feature reuse &62.3 &64.9&68.7 &\textbf{-1.6}\\  
				3 &$w/$ Residual &62.2 &64.7&68.9 &\textbf{-1.5}\\  
				4 &$w/$ Concatenation &62.0 &65.4&69.0 &\textbf{-1.4}\\ 
				5 &$w/$ Gated summation &64.5 &66.6&69.3 &\textbf{-0.1}\\  
				\hdashline
				6 &$w/o$ Target-aware self-distillation &62.9 &65.1 &69.4 &\textbf{-1.2}\\
				7 &only self-distillation &63.8 &66.7 &69.1 &\textbf{-0.3}\\
				\hdashline
				8 &Joint $\rightarrow$ Fixed  Backbone &62.5 &65.5&69.0 &\textbf{-1.2}\\
				9 &Joint $\rightarrow$ One-by-one &62.3 &66.2&69.0 &\textbf{-1.1}\\
				\hdashline
				10 & Baseline (separately trained) &61.1 &64.7 &68.4 &\textbf{-2.2}\\
				\hline
			\end{tabular}
	}}
	\vspace{-0mm}
	\caption{Exploration studies on LaSOT~\cite{lasot}. 
		The performance (AUC score) from early to deep exits is denoted by E-1, E-2, and E-3.
		$\Delta$ denotes the performance change (averaged over exits) compared with the baseline. \#10 is the baseline methods with the same layer numbers as different exits.
	}
	\label{tab:ablation}
	\vspace{-0mm}
\end{table}

\subsection{Exploration Studies}
\label{sec:ablation}
In this section, we inspect different aspects of DyTrack.
Unless stated otherwise, we make the models predict at each single exiting node and analyze the results from all nodes for a fair comparison. For all ablations, we use the large-scale benchmark LaSOT to report the performance.

\noindent\textbf{Design choices for feature reuse.}
We introduce a feature recycling mechanism that reduces the waste and redundancy of computations by feature reuse.
Hereby, we compare four variants of reuse manner 
as described in Section~\ref{sec:FR}.
The corresponding exploratory experiments are shown in Table~\ref{tab:ablation}.
As shown in experiment \#2, without feature reuse, the performance declines significantly, especially at the early and intermediate exits.
It indicates that computations from early decisioner networks can be exploited to facilitate better box prediction in our dynamic transformer structure.
Experiments \#3, \#1, \#4, and \#5 in Table~\ref{tab:ablation} represent employing the feature reuse schemas (a), (b), (c), and (d) in Figure~\ref{fig:feat_reuse}.
We can find that residual integration can not bring improvements.
This may be because the decisioner network recieves the input features from the encoder, so it is more suitable for learning direct representations of the target rather than its residuals.
And applying element-wise summation to the input features allows the recycled features to undergo deeper representations, thereby providing more discriminative cues.
The concatenation operation also proves to be slightly beneficial. Its performance is inferior to that of (b), primarily because (b) leverages the spatial alignment between recycled features and current features, enhancing critical spatial information while avoiding the additional computational cost introduced by the increased number of tokens after concatenation. 
We observe that the gated summation scheme brings improvements compared to \#2. However, it does not achieve promising performance gains compared to schema (b), likely because the recycled features exhibit no significant redundancy. Moreover, the additional computations introduced by the gate operation reduce the efficiency. 
Therefore, we adopt schema (b), which applies a summation operation before sending features into the Decisioner networks. This simple operation is found to deliver the best performance.

\begin{figure}[t]
	\begin{center}
		\includegraphics[width=0.98\linewidth]{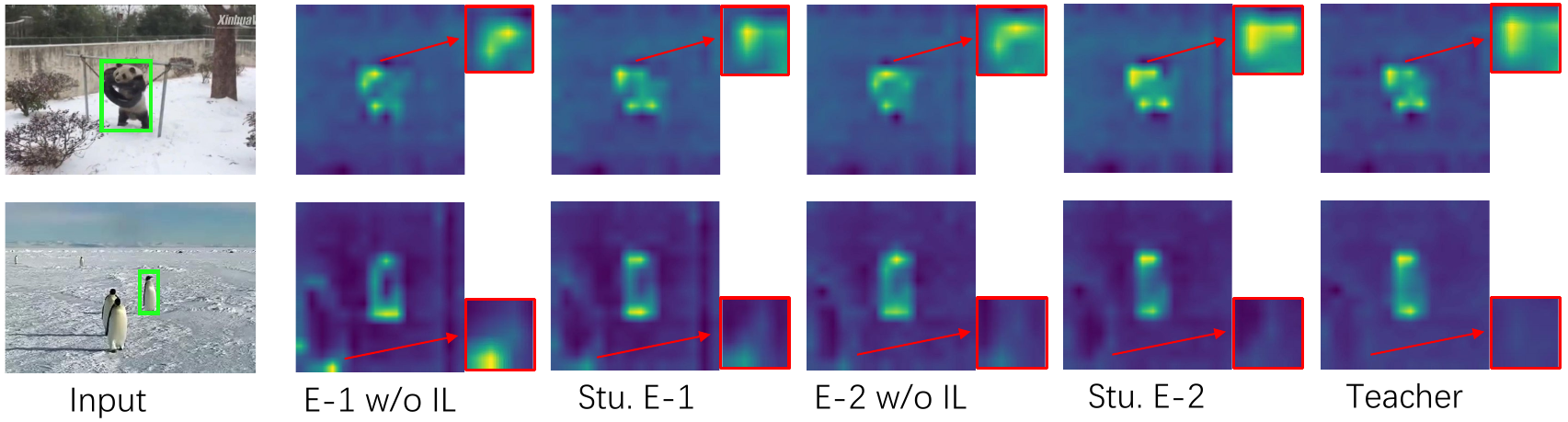}
	\end{center}
	\vspace{-3mm}
	\caption{
		Qualitative results of with and without the proposed target-aware self-distillation.
	}
	\vspace{-2mm}
	\label{vis_imitation}
\end{figure}

\begin{figure}[t]
	\centering
	\vspace{-2mm}
	\includegraphics[width=0.475\textwidth]{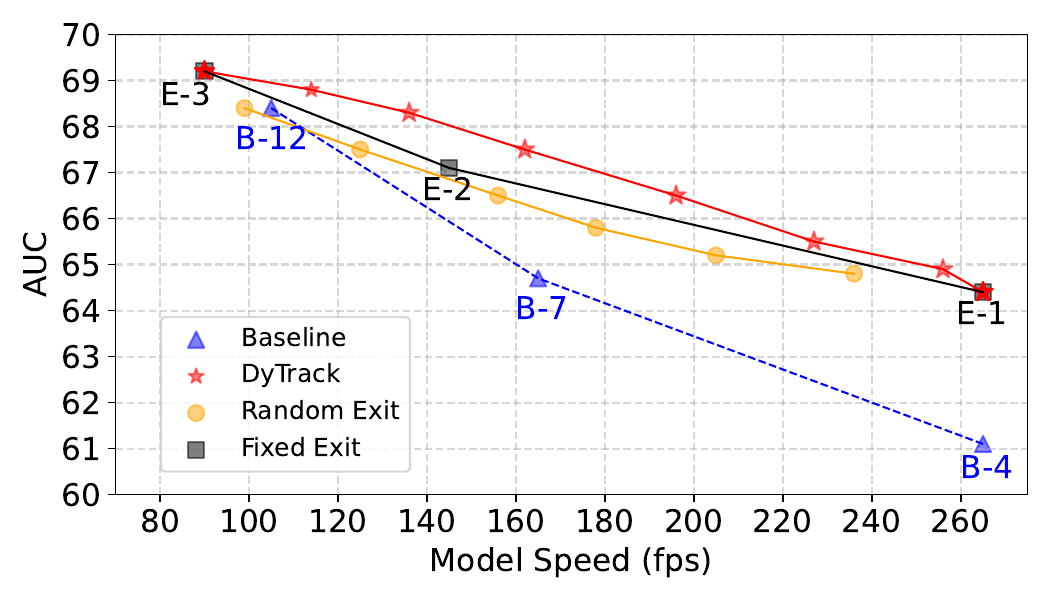}
	\caption{Ablation studies of different exiting rules. The solid lines indicate results from the same model, the dashed lines indicate results from different models.}
	\vspace{-2mm}
	\label{fig:ao_fps_ablation}
\end{figure}

\noindent\textbf{Impact of target-aware self-distillation.}
As described in Section~\ref{sec:tasd}, 
target-aware self-distillation is designed for enhancing the feature modeling of the early branches by executing auxiliary supervision from the deep layer.
We validate the impact of the proposed strategy in Table~\ref{tab:ablation} (\#6 and \#7).
We can see that self-distillation (removing imitation attention) improves the performance of early and intermediate exits (\#7), which can be further facilitated by target-aware imitation loss (\#6).
The experiments show that the dynamic tracking model can improve the performance of all early routes through a self-distillation mechanism without adding additional inference burdens.
Furthermore, we showcase the generated feature maps from the teacher and the early nodes with and without imitation learning to intuitively verify the validity of the imitation learning mechanism.
As shown in Figure~\ref{vis_imitation}, trained with feature imitation learning (IL), the activation maps from early nodes are closer to the teacher’s.

\noindent\textbf{Different training strategies.}
As described in Section~\ref{sec:implem_detail}, 
we joint train a pre-trained base model with the decisioner networks and prediction heads in the second stage.
Here, we compare our training strategy with two others.
These training strategies both train a base model, but then
\textit{\textbf{i})} fix the backbone of the learned base model, 
and only train the exiting branches at multiple nodes (see Table~\ref{tab:ablation} \#8).
\textit{\textbf{ii})} fix the backbone,
and train the exiting branches one by one, from early to deep layers. Every single training for an exiting branch takes 120 epoch with the same other training settings (as shown in Table~\ref{tab:ablation} \#9).
The results show that joint training all the exiting branches along with the backbone get the best performance. The underlying reason might be that the joint training allows the entire model to be supervised at multiple levels at the same time, facilitating learning towards the overall optimum.

\begin{figure}[b]
	\centering
	\begin{subfloat}
		\centering   
		\includegraphics[width=0.64\linewidth]{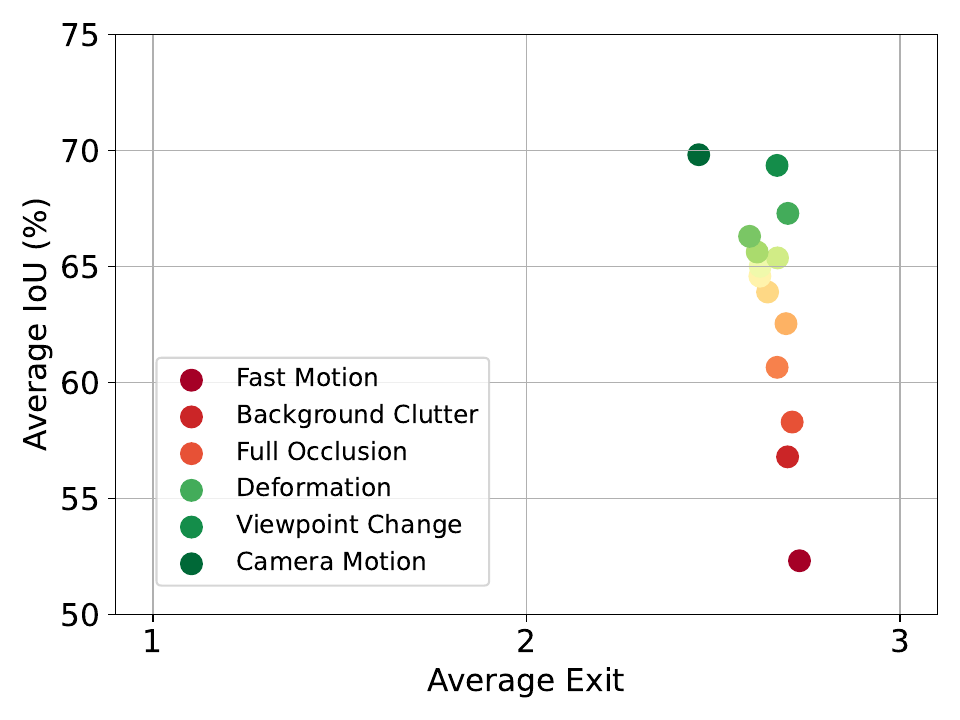}
	\end{subfloat}  
	\hspace{-4mm}
	\begin{subfloat}
		\centering
		\includegraphics[width=0.34\linewidth]{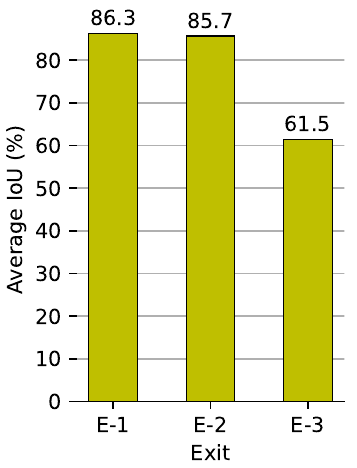}
	\end{subfloat}
	\vspace{-1mm}
	\caption{Exploration studies on video attributes and exits. Left: average IoU scores $vs$ average exits for videos of different attributes. Right: average IoU scores $vs$ different exits.}
	\label{fig:attr_iou_exit}
	\vspace{-0mm}
\end{figure}

\noindent\textbf{Fixed $vs$ adaptive exiting.}
DyTrack performs instance-specific inference so that samples of different complexities are assigned to different routes to return the prediction.
Here, we make DyTrack to output results at the fixed nodes and compare them with 
adaptive exiting in Figure~\ref{fig:ao_fps_ablation} (Fixed Exit $vs$ DyTrack).
Adaptive exiting demonstrates superiority in both speed and accuracy, due to the fact that DyTrack transfers the computations on easy samples to the processing of complex ones, thus improving both effectiveness and accuracy.
When DyTrack degenerates to the most complex and simplest state, all samples will pass through the final and the first node for output prediction, respectively.

\noindent\textbf{Early-termination criterion.}
We embed an IoU token in the encoder, which is used in the decisioner network to predict the tracking score of the current frame, and this is used as a criterion for whether DyTrack should perform an early termination.
We compare the IoU score criterion with random exiting to verify the validity of the exiting conditions.
As shown in Figure~\ref{fig:ao_fps_ablation}, random exiting shows worse speed-precision trade-offs, also below the solution of fixed exit.
The reason may be that random exiting causes the model to make some inefficient assignments, \eg
assigning simple routes to complex samples and complex routes to simple inputs, resulting in lower precision and worse efficiency.
This experiment also demonstrates the effectiveness of using the IoU score as a termination criterion.

\begin{figure}[t]
	\centering
	\includegraphics[width=0.46\textwidth]{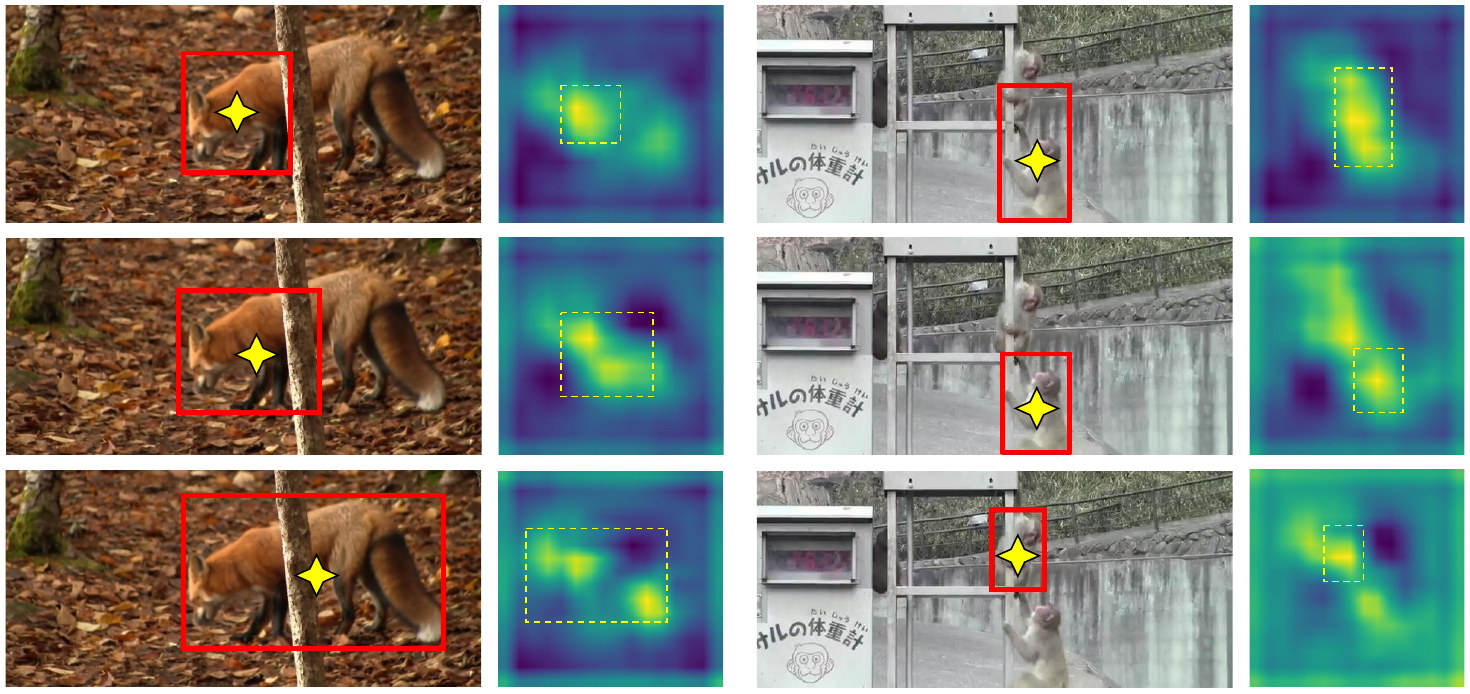}
	\vspace{-0mm}
	\caption{Visual features from different exits. The first to last rows indicate the predicted results from the first to last exits. More complex reasoning routes lead to more accurate tracking.}
	\vspace{-0mm}
	\label{fig:vis}
\end{figure}

\noindent\textbf{Influence of attributes and exits.}
\label{sec:exit_analysis}
We control the inference condition that the ratio of output frames by all three exits to $1:1:1$ to study the influence of attributes and prediction exits.
Figure~\ref{fig:attr_iou_exit} (left) illustrates the average precision of all attributes in the LaSOT~\cite{lasot} benchmark as well as the average precision for each exit category. 
As can be seen, the difficult cases (with attributes such as \textit{Fast Motion}, \textit{Background Clutter}, \textit{and Full Occlusion}) require complex routes to handle.
In contrast, some videos with attributes such as \textit{Camera Motion}, \textit{Viewpoint Change}, and \textit{Deformation} appear to be simpler for DyTrack and therefore exit in more early exits with high accuracy. 
Figure~\ref{fig:attr_iou_exit} (right) shows another interesting observation that the more we execute exiting in later stages, the more inaccurate the predictions become. 
This may sound counter-intuitive because if we get a model without instance-specific reasoning, late-stage exits produce the most accurate predictions.
However, the trend shown in Figure~\ref{fig:attr_iou_exit} (right) is desirable, because 
the easier frames have already exited from the network while only hard ones reach late-stage routes.

\begin{figure*}[tp]
	\centering
	\vspace{0mm}
	\includegraphics[width=0.95\linewidth]{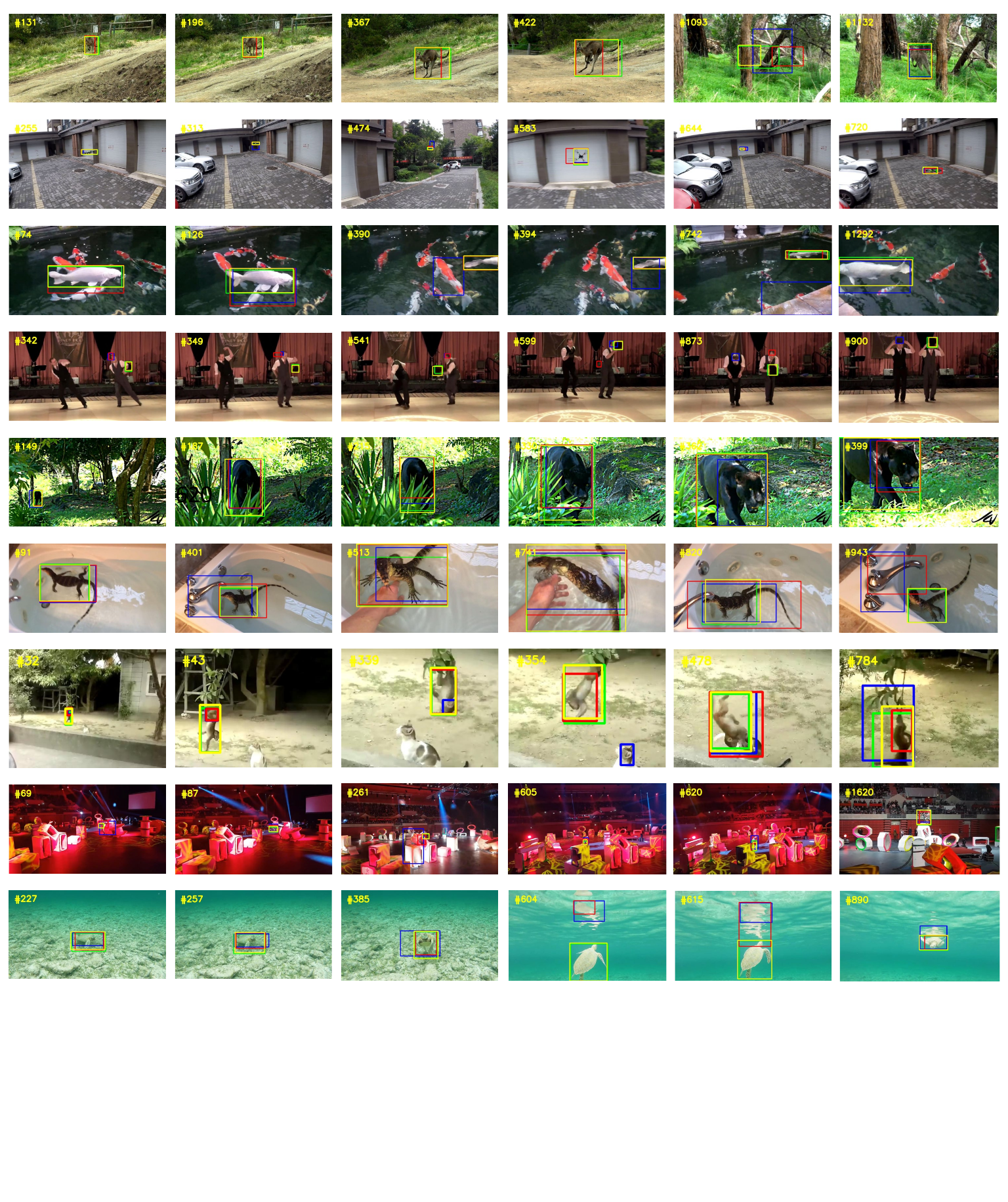}
	\caption{Tracking results from different exits. The \textcolor{green}{green} bounding boxes denote ground truth, and the \textcolor{blue}{blue}, \textcolor{red}{red}, and \textcolor{yellow}{yellow} bounding boxes denote DyTrack-Fast, DyTrack-Medi, and DyTrack-Base, respectively. Best viewed in color with zoom-in.}
	\label{fig:vis_exits}
	\vspace{0mm}
\end{figure*}

\noindent\textbf{Visualization of various exits.}
We visualize the combination score map (top-left + bottom-right corners) to understand the different representational capabilities of different reasoning routes.
As shown in Figure~\ref{fig:vis},
when dealing with some complex instances, \eg occlusion and distractor, 
simple inference route struggles to perform, and as the complexity of the network increases, attention is more focused on the target object.
Different reasoning routes in dynamic networks show clear capability gaps.
Figure~\ref{fig:vis_exits} showcases more qualitative results of DyTrack in different tracking scenarios including deformation, background clutter, \etc

\begin{figure}[t]
	\centering
	\includegraphics[width=1\linewidth]{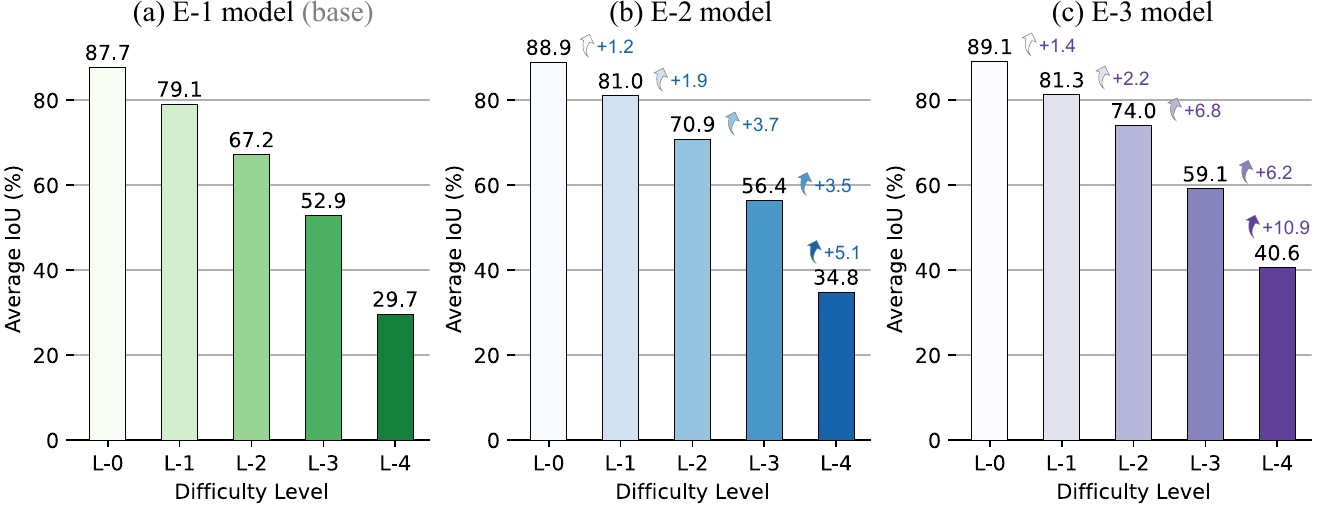}
	\vspace{-5mm}
	\caption{Evaluation of models of different sizes on test subsets with varying difficulty levels.}
	\label{fig: iou_vs_model}
\end{figure}

\noindent\textbf{Performance analysis on varying difficulty levels.}
To further analyze the effect of model size on different difficulty levels, we divide the test sequences (from the LaSOT benchmark) into 5 difficulty levels using a small model.
The small model is the E-1 model of DyTrack, processes inputs through the simplest path and directly outputs predictions from the 1st exit. By sorting the predicted results' IoU scores across all sequences, we uniformly divide the sequences into 5 subsets (each containing 56 sequences), representing 5 difficulty levels.
The IoU scores of the E-1 model on the test subsets are shown in Fig.~\ref{fig: iou_vs_model} (a). For sequences with the lowest difficulty level (L-0), the smallest E-1 model achieves an average IoU of 87.7\%. However, as the difficulty level increases, the IoU scores start to decline, reaching only 29.7\% on the most challenging sequences at the L-4 difficulty level.
When using larger (i.e., more complex) models, the performance consistently improves across all difficulty levels. A noticeable trend is that larger models yield significantly greater improvements on subsets with higher difficulty levels. For instance, the E-2 model achieves only a 1.2\% IoU improvement over the base E-1 model on L-0 sequences but demonstrates a substantial 5.1\% improvement on the most challenging L-4 sequences. When the model is replaced with the even larger E-3 model, the IoU score on L-4 sequences improves by a remarkable 10.9\%, clearly demonstrating that larger models are indeed better at handling difficult cases.

\begin{figure}[h]
	\centering
	\includegraphics[width=0.98\linewidth]{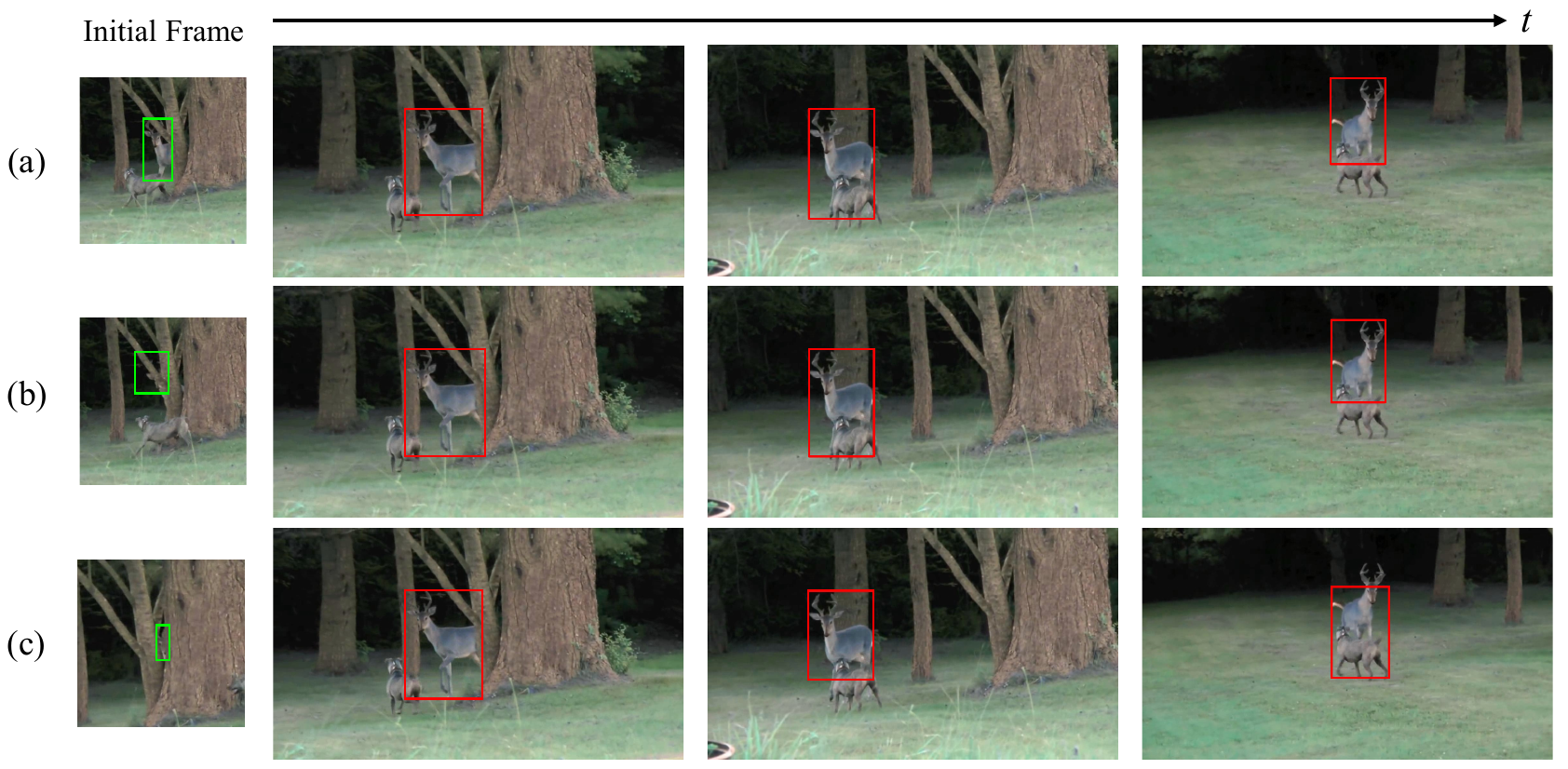}
	\vspace{-1mm}
	\caption{Initialize DyTrack with partially visible objects.}
	\label{fig: occ_init}
\end{figure}

\noindent\textbf{Initializing with partially visible objects.}
We test DyTrack under varying degrees of target visibility in the start frame. The results are presented Fig.~\ref{fig: occ_init}.
In Figure~\ref{fig: occ_init} (a), the target deer exhibits relatively clear and visible features, such as its head and the front part of its body. Consequently, DyTrack can accurately and stably track the target in subsequent frames.
In Figure~\ref{fig: occ_init} (b), we further reduce the visible area, showing only part of the deer’s head and antlers in the start frame. DyTrack demonstrates strong robustness, successfully tracking the target in subsequent frames. 
This indicates that DyTrack can extract effective target cues from the start frame, even when a limited portion of the target is visible.
To increase the difficulty further, we tested a scenario where only a small portion of the deer’s face is visible in the initial frame, and the target exhibits camouflage effects due to its similar color and texture to the tree trunk (Figure~\ref{fig: occ_init} (c)). Surprisingly, DyTrack is still able to track the target in most cases, especially in the early time of the video sequence.
Initializing with a partially visible object poses a challenge for trackers. In the future, DyTrack could further enhance its robustness by incorporating techniques such as multiple template and template updating.

\begin{table}[t]
	\small
	\centering
	\renewcommand\arraystretch{1.05}
	\resizebox{\linewidth}{!}{
		\setlength{\tabcolsep}{0.5mm}{
			\begin{tabular}{c|ccccc|ccc}
				\hline
				\small
				Method&ECO&ATOM&DiMP&TransT&MixFormer&\tabincell{c}{{DyTrack}\\{-Fast}}&\tabincell{c}{{DyTrack}\\{-Medi}}&\tabincell{c}{{DyTrack}\\{-Base}}\\
				\hline
				AUC&32.4&51.5&56.9&64.9&69.2&64.9&66.5&69.2 \\
				$fps$&39&22&17&13&-&37&27&11 \\
				\hline
		\end{tabular}}
	}
	\vspace{-0mm}
	\caption{Speed $vs$ performance comparison on AGX.}
	\vspace{-0mm}
	\label{tab-agx}
\end{table}

\noindent\textbf{Efficiency on edge device.}
At last, we present an additional
experiment to demonstrate that DyTrack is flexible to adapt to edge computing platform. 
We test our method on an Nvidia Jetson AGX Xavier (AGX) which can reflect the model efficiency in resource-constrained devices.
As shown in Table~\ref{tab-agx}, by configuring faster reasoning routes, DyTrack-Fast and DyTrack-Medi can meet the real-time requirement (25$fps$) on AGX. Meanwhile, with similar speeds, DyTrack outperforms other methods by a large margin.

\section{Limitations}
One limitation of DyTrack is that, despite achieving competitive results with promising speed-precision trade-offs, our research of dynamic network mainly focuses on the depth of the network, which may not be optimal.
Combining dynamic spatial feature selecting might further improve the efficiency.

\section{Conclusion}

This work proposes DyTrack, a dynamic transformer tracking framework for efficient object tracking.
Different with existing efficient tracking methods that primarily focus on designing lightweight modules, DyTrack works with an instance-specific reasoning fashion that learns to assign various inference routes for different frames, reducing the computations for easy scenarios. 
We further introduce feature recycling and target-aware self-distillation to efficiency and precision, respectively. Extensive experiments demonstrate that DyTrack achieves a superior speed-precision trade-off compared to both high-performance and high-speed methods.

\bibliographystyle{IEEEtran}
\bibliography{egbib}

\begin{thebibliography}{10}
\providecommand{\url}[1]{#1}
\csname url@samestyle\endcsname
\providecommand{\newblock}{\relax}
\providecommand{\bibinfo}[2]{#2}
\providecommand{\BIBentrySTDinterwordspacing}{\spaceskip=0pt\relax}
\providecommand{\BIBentryALTinterwordstretchfactor}{4}
\providecommand{\BIBentryALTinterwordspacing}{\spaceskip=\fontdimen2\font plus
\BIBentryALTinterwordstretchfactor\fontdimen3\font minus
  \fontdimen4\font\relax}
\providecommand{\BIBforeignlanguage}[2]{{%
\expandafter\ifx\csname l@#1\endcsname\relax
\typeout{** WARNING: IEEEtran.bst: No hyphenation pattern has been}%
\typeout{** loaded for the language `#1'. Using the pattern for}%
\typeout{** the default language instead.}%
\else
\language=\csname l@#1\endcsname
\fi
#2}}
\providecommand{\BIBdecl}{\relax}
\BIBdecl

\bibitem{siameserpn}
B.~Li, J.~Yan, W.~Wu, Z.~Zhu, and X.~Hu, ``High performance visual tracking
  with siamese region proposal network,'' in \emph{Proceedings of the IEEE/CVF
  Conference on Computer Vision and Pattern Recognition}, 2018, pp. 8971--8980.

\bibitem{yu2022learning}
H.~Yu, P.~Zhu, K.~Zhang, Y.~Wang, S.~Zhao, L.~Wang, T.~Zhang, and Q.~Hu,
  ``Learning dynamic compact memory embedding for deformable visual object
  tracking,'' \emph{IEEE Transactions on Neural Networks and Learning Systems},
  vol.~35, no.~4, pp. 5656--5670, 2022.

\bibitem{transt}
X.~Chen, B.~Yan, J.~Zhu, D.~Wang, X.~Yang, and H.~Lu, ``Transformer tracking,''
  in \emph{Proceedings of the IEEE/CVF Conference on Computer Vision and
  Pattern Recognition}, 2021, pp. 8126--8135.

\bibitem{srrt}
J.~Zhu, X.~Chen, P.~Zhang, X.~Wang, D.~Wang, W.~Zhao, and H.~Lu, ``{SRRT}:
  Exploring search region regulation for visual object tracking,'' \emph{IEEE
  Transactions on Circuits and Systems for Video Technology}, vol.~34, no.~11,
  pp. 10\,551--10\,563, 2024.

\bibitem{ostrack}
B.~Ye, H.~Chang, B.~Ma, S.~Shan, and X.~Chen, ``Joint feature learning and
  relation modeling for tracking: A one-stream framework,'' in \emph{European
  Conference on Computer Vision}, 2022, pp. 341--357.

\bibitem{mixformer}
Y.~Cui, C.~Jiang, L.~Wang, and G.~Wu, ``{Mixformer}: End-to-end tracking with
  iterative mixed attention,'' in \emph{Proceedings of the IEEE/CVF Conference
  on Computer Vision and Pattern Recognition}, 2022, pp. 13\,608--13\,618.

\bibitem{alexnet}
A.~Krizhevsky, I.~Sutskever, and G.~E. Hinton, ``Imagenet classification with
  deep convolutional neural networks,'' in \emph{Advances in Neural Information
  Processing Systems}, 2012, pp. 1097--1105.

\bibitem{resnet}
K.~He, X.~Zhang, S.~Ren, and J.~Sun, ``Deep residual learning for image
  recognition,'' in \emph{Proceedings of the IEEE/CVF Conference on Computer
  Vision and Pattern Recognition}, 2016, pp. 770--778.

\bibitem{vit}
A.~Dosovitskiy, L.~Beyer, A.~Kolesnikov, D.~Weissenborn, X.~Zhai,
  T.~Unterthiner, M.~Dehghani, M.~Minderer, G.~Heigold, S.~Gelly, J.~Uszkoreit,
  and N.~Houlsby, ``An image is worth 16x16 words: Transformers for image
  recognition at scale,'' in \emph{International Conference on Learning
  Representations}, 2021, pp. 1--21.

\bibitem{attention_is_all}
A.~Vaswani, N.~Shazeer, N.~Parmar, J.~Uszkoreit, L.~Jones, A.~N. Gomez,
  {\L}.~Kaiser, and I.~Polosukhin, ``Attention is all you need,'' in
  \emph{Advances in Neural Information Processing Systems}, 2017, pp.
  5998--6008.

\bibitem{eco}
M.~Danelljan, G.~Bhat, F.~Shahbaz~Khan, and M.~Felsberg, ``{ECO}: {E}fficient
  convolution operators for tracking,'' in \emph{Proceedings of the IEEE/CVF
  Conference on Computer Vision and Pattern Recognition}, Honolulu, HI, USA,
  2017, pp. 6638--6646.

\bibitem{atom}
M.~Danelljan, G.~Bhat, F.~S. Khan, and M.~Felsberg, ``{ATOM}: {A}ccurate
  tracking by overlap maximization,'' in \emph{Proceedings of the IEEE/CVF
  Conference on Computer Vision and Pattern Recognition}, 2019, pp. 4660--4669.

\bibitem{lighttrack}
B.~Yan, H.~Peng, K.~Wu, D.~Wang, J.~Fu, and H.~Lu, ``Lighttrack: Finding
  lightweight neural networks for object tracking via one-shot architecture
  search,'' in \emph{Proceedings of the IEEE/CVF Conference on Computer Vision
  and Pattern Recognition}, 2021, pp. 15\,180--15\,189.

\bibitem{fear}
V.~Borsuk, R.~Vei, O.~Kupyn, T.~Martyniuk, I.~Krashenyi, and J.~Matas,
  ``{FEAR}: {F}ast, efficient, accurate and robust visual tracker,'' in
  \emph{European Conference on Computer Vision}.\hskip 1em plus 0.5em minus
  0.4em\relax Springer, 2022, pp. 644--663.

\bibitem{asymtrack}
J.~Zhu, H.~Tang, X.~Chen, X.~Wang, D.~Wang, and H.~Lu, ``Two-stream beats
  one-stream: Asymmetric siamese network for efficient visual tracking,''
  \emph{arXiv preprint arXiv:2503.00516}, 2025.

\bibitem{dcf}
A.~Lukezic, T.~Vojir, L.~{\v{C}}ehovin~Zajc, J.~Matas, and M.~Kristan,
  ``Discriminative correlation filter with channel and spatial reliability,''
  in \emph{Proceedings of the IEEE/CVF Conference on Computer Vision and
  Pattern Recognition}, 2017, pp. 6309--6318.

\bibitem{lasot}
H.~Fan, L.~Lin, F.~Yang, P.~Chu, G.~Deng, S.~Yu, H.~Bai, Y.~Xu, C.~Liao, and
  H.~Ling, ``{LaSOT}: A high-quality benchmark for large-scale single object
  tracking,'' in \emph{Proceedings of the IEEE/CVF Conference on Computer
  Vision and Pattern Recognition}, 2019, pp. 5374--5383.

\bibitem{pmn}
X.~Liang, ``Learning personalized modular network guided by structured
  knowledge,'' in \emph{Proceedings of the IEEE/CVF Conference on Computer
  Vision and Pattern Recognition}, 2019, pp. 8944--8952.

\bibitem{kaya2019shallow}
Y.~Kaya, S.~Hong, and T.~Dumitras, ``Shallow-deep networks: Understanding and
  mitigating network overthinking,'' in \emph{International Conference on
  Machine Learning}.\hskip 1em plus 0.5em minus 0.4em\relax PMLR, 2019, pp.
  3301--3310.

\bibitem{cai2021dynamic}
S.~Cai, Y.~Shu, and W.~Wang, ``Dynamic routing networks,'' in \emph{Proceedings
  of the IEEE/CVF Winter Conference on Applications of Computer Vision}, 2021,
  pp. 3588--3597.

\bibitem{pbee}
W.~Zhou, C.~Xu, T.~Ge, J.~McAuley, K.~Xu, and F.~Wei, ``Bert loses patience:
  Fast and robust inference with early exit,'' \emph{Advances in Neural
  Information Processing Systems}, vol.~33, pp. 18\,330--18\,341, 2020.

\bibitem{sun2021early}
T.~Sun, Y.~Zhou, X.~Liu, X.~Zhang, H.~Jiang, Z.~Cao, X.~Huang, and X.~Qiu,
  ``Early exiting with ensemble internal classifiers,'' \emph{arXiv preprint
  arXiv:2105.13792}, 2021.

\bibitem{huang2017multi}
G.~Huang, D.~Chen, T.~Li, F.~Wu, L.~Van Der~Maaten, and K.~Q. Weinberger,
  ``Multi-scale dense networks for resource efficient image classification,''
  in \emph{International Conference on Learning Representations}, 2018, pp.
  1--14.

\bibitem{siamesefc}
L.~Bertinetto, J.~Valmadre, J.~F. Henriques, A.~Vedaldi, and P.~H. Torr,
  ``Fully-convolutional siamese networks for object tracking,'' in
  \emph{European Conference on Computer Vision Workshops}, 2016, pp. 850--865.

\bibitem{fast_rcnn}
R.~Girshick, ``Fast r-cnn,'' in \emph{Proceedings of the IEEE international
  Conference on Computer Vision}, 2015, pp. 1440--1448.

\bibitem{imagenet}
O.~Russakovsky, J.~Deng, H.~Su, J.~Krause, S.~Satheesh, S.~Ma, Z.~Huang,
  A.~Karpathy, A.~Khosla, M.~Bernstein \emph{et~al.}, ``Imagenet large scale
  visual recognition challenge,'' \emph{International Journal of Computer
  Vision}, vol. 115, no.~3, pp. 211--252, 2015.

\bibitem{youtube-vos}
N.~Xu, L.~Yang, Y.~Fan, J.~Yang, D.~Yue, Y.~Liang, B.~Price, S.~Cohen, and
  T.~Huang, ``Youtube-vos: Sequence-to-sequence video object segmentation,'' in
  \emph{European Conference on Computer Vision}, 2018, pp. 585--601.

\bibitem{siamrpnplusplus}
B.~Li, W.~Wu, Q.~Wang, F.~Zhang, J.~Xing, and J.~Yan, ``{SiamRPN++}:
  {E}volution of siamese visual tracking with very deep networks,'' in
  \emph{Proceedings of the IEEE/CVF Conference on Computer Vision and Pattern
  Recognition}, 2019, pp. 4282--4291.

\bibitem{ma2017robust}
B.~Ma, H.~Hu, J.~Shen, Y.~Zhang, L.~Shao, and F.~Porikli, ``Robust object
  tracking by nonlinear learning,'' \emph{IEEE Transactions on Neural Networks
  and Learning Systems}, vol.~29, no.~10, pp. 4769--4781, 2017.

\bibitem{dimp}
G.~Bhat, M.~Danelljan, L.~V. Gool, and R.~Timofte, ``Learning discriminative
  model prediction for tracking,'' in \emph{Proceedings of the IEEE
  International Conference on Computer Vision}, 2019, pp. 6182--6191.

\bibitem{zhang2023attention}
H.~Zhang, J.~Liang, J.~Zhang, T.~Zhang, Y.~Lin, and Y.~Wang, ``Attention-driven
  memory network for online visual tracking,'' \emph{IEEE Transactions on
  Neural Networks and Learning Systems}, vol.~35, no.~12, pp. 17\,085--17\,098,
  2023.

\bibitem{swintrack}
L.~Lin, H.~Fan, Z.~Zhang, Y.~Xu, and H.~Ling, ``{SwinTrack}: A simple and
  strong baseline for transformer tracking,'' in \emph{Advances in Neural
  Information Processing Systems}, 2022, pp. 16\,743--16\,754.

\bibitem{grm}
S.~Gao, C.~Zhou, and J.~Zhang, ``Generalized relation modeling for transformer
  tracking,'' in \emph{Proceedings of the IEEE/CVF Conference on Computer
  Vision and Pattern Recognition}, 2023, pp. 18\,686--18\,695.

\bibitem{swin}
Z.~Liu, Y.~Lin, Y.~Cao, H.~Hu, Y.~Wei, Z.~Zhang, S.~Lin, and B.~Guo, ``Swin
  transformer: Hierarchical vision transformer using shifted windows,'' in
  \emph{Proceedings of the IEEE International Conference on Computer Vision},
  2021, pp. 10\,012--10\,022.

\bibitem{simtrack}
B.~Chen, P.~Li, L.~Bai, L.~Qiao, Q.~Shen, B.~Li, W.~Gan, W.~Wu, and W.~Ouyang,
  ``Backbone is all your need: a simplified architecture for visual object
  tracking,'' in \emph{European Conference on Computer Vision}.\hskip 1em plus
  0.5em minus 0.4em\relax Springer, 2022, pp. 375--392.

\bibitem{vipt}
J.~Zhu, S.~Lai, X.~Chen, D.~Wang, and H.~Lu, ``Visual prompt multi-modal
  tracking,'' in \emph{Proceedings of the IEEE/CVF Conference on Computer
  Vision and Pattern Recognition}, 2023, pp. 9516--9526.

\bibitem{cvt}
H.~Wu, B.~Xiao, N.~Codella, M.~Liu, X.~Dai, L.~Yuan, and L.~Zhang, ``{CvT}:
  Introducing convolutions to vision transformers,'' in \emph{Proceedings of
  the IEEE International Conference on Computer Vision}, 2021, pp. 22--31.

\bibitem{mae}
K.~He, X.~Chen, S.~Xie, Y.~Li, P.~Doll{\'a}r, and R.~Girshick, ``Masked
  autoencoders are scalable vision learners,'' in \emph{Proceedings of the
  IEEE/CVF Conference on Computer Vision and Pattern Recognition}, 2022, pp.
  16\,000--16\,009.

\bibitem{li2023self}
X.~Li, W.~Pei, Y.~Wang, Z.~He, H.~Lu, and M.-H. Yang, ``Self-supervised
  tracking via target-aware data synthesis,'' \emph{IEEE Transactions on Neural
  Networks and Learning Systems}, vol.~35, no.~7, pp. 9186--9197, 2023.

\bibitem{pham2018efficient}
H.~Pham, M.~Guan, B.~Zoph, Q.~Le, and J.~Dean, ``Efficient neural architecture
  search via parameters sharing,'' in \emph{International Conference on Machine
  Learning}.\hskip 1em plus 0.5em minus 0.4em\relax PMLR, 2018, pp. 4095--4104.

\bibitem{shufflenetv2}
N.~Ma, X.~Zhang, H.-T. Zheng, and J.~Sun, ``Shufflenet v2: Practical guidelines
  for efficient cnn architecture design,'' in \emph{European Conference on
  Computer Vision}, 2018, pp. 116--131.

\bibitem{ussa2023hybrid}
A.~Ussa, C.~S. Rajen, T.~Pulluri, D.~Singla, J.~Acharya, G.~F. Chuanrong,
  A.~Basu, and B.~Ramesh, ``A hybrid neuromorphic object tracking and
  classification framework for real-time systems,'' \emph{IEEE Transactions on
  Neural Networks and Learning Systems}, vol.~35, no.~8, pp. 10\,726--10\,735,
  2023.

\bibitem{ettrack}
P.~Blatter, M.~Kanakis, M.~Danelljan, and L.~Van~Gool, ``Efficient visual
  tracking with exemplar transformers,'' in \emph{IEEE Winter Conference on
  Applications of Computer Vision}, 2023, pp. 1571--1581.

\bibitem{litetrack}
Q.~Wei, B.~Zeng, J.~Liu, L.~He, and G.~Zeng, ``{LiteTrack}: {Layer} pruning
  with asynchronous feature extraction for lightweight and efficient visual
  tracking,'' in \emph{IEEE International Conference on Robotics and
  Automation}.\hskip 1em plus 0.5em minus 0.4em\relax IEEE, 2024, pp.
  4968--4975.

\bibitem{mixformerv2}
Y.~Cui, T.~Song, G.~Wu, and L.~Wang, ``{MixFormerV2}: {Efficient} fully
  transformer tracking,'' in \emph{Advances in Neural Information Processing
  Systems}, 2023, pp. 58\,736--58\,751.

\bibitem{hit}
B.~Kang, X.~Chen, D.~Wang, H.~Peng, and H.~Lu, ``Exploring lightweight
  hierarchical vision transformers for efficient visual tracking,'' in
  \emph{Proceedings of the IEEE/CVF International Conference on Computer
  Vision}, 2023, pp. 9612--9621.

\bibitem{abtrack}
X.~Yang, D.~Zeng, X.~Wang, Y.~Wu, H.~Ye, Q.~Zhao, and S.~Li, ``Adaptively
  bypassing vision transformer blocks for efficient visual tracking,''
  \emph{arXiv preprint arXiv:2406.08037}, 2024.

\bibitem{zhang2023efficient}
T.~Zhang, H.~Guo, Q.~Jiao, Q.~Zhang, and J.~Han, ``Efficient {RGB-T} tracking
  via cross-modality distillation,'' in \emph{Proceedings of the IEEE/CVF
  Conference on Computer Vision and Pattern Recognition}, 2023, pp. 5404--5413.

\bibitem{east}
C.~Huang, S.~Lucey, and D.~Ramanan, ``Learning policies for adaptive tracking
  with deep feature cascades,'' in \emph{Proceedings of the IEEE International
  Conference on Computer Vision}, 2017, pp. 105--114.

\bibitem{skipnet}
X.~Wang, F.~Yu, Z.-Y. Dou, T.~Darrell, and J.~E. Gonzalez, ``{SkipNet}:
  {Learning} dynamic routing in convolutional networks,'' in \emph{European
  Conference on Computer Vision}, Munich, Germany, 2018, pp. 409--424.

\bibitem{veit2018convolutional}
A.~Veit and S.~Belongie, ``Convolutional networks with adaptive inference
  graphs,'' in \emph{European Conference on Computer Vision}, 2018, pp. 3--18.

\bibitem{blockdrop}
Z.~Wu, T.~Nagarajan, A.~Kumar, S.~Rennie, L.~S. Davis, K.~Grauman, and
  R.~Feris, ``Blockdrop: Dynamic inference paths in residual networks,'' in
  \emph{Proceedings of the IEEE/CVF Conference on Computer Vision and Pattern
  Recognition}, 2018, pp. 8817--8826.

\bibitem{dvt}
Y.~Wang, R.~Huang, S.~Song, Z.~Huang, and G.~Huang, ``Not all images are worth
  16x16 words: Dynamic transformers for efficient image recognition,''
  \emph{Advances in Neural Information Processing Systems}, vol.~34, pp.
  11\,960--11\,973, 2021.

\bibitem{ztw}
M.~Wo{\l}czyk, B.~W{\'o}jcik, K.~Ba{\l}azy, I.~T. Podolak, J.~Tabor,
  M.~{\'S}mieja, and T.~Trzcinski, ``Zero time waste: Recycling predictions in
  early exit neural networks,'' \emph{Advances in Neural Information Processing
  Systems}, vol.~34, pp. 2516--2528, 2021.

\bibitem{gaternet}
Z.~Chen, Y.~Li, S.~Bengio, and S.~Si, ``Gaternet: Dynamic filter selection in
  convolutional neural network via a dedicated global gating network,''
  \emph{arXiv preprint arXiv:1811.11205}, 2018.

\bibitem{abati2020conditional}
D.~Abati, J.~Tomczak, T.~Blankevoort, S.~Calderara, R.~Cucchiara, and B.~E.
  Bejnordi, ``Conditional channel gated networks for task-aware continual
  learning,'' in \emph{Proceedings of the IEEE/CVF Conference on Computer
  Vision and Pattern Recognition}, 2020, pp. 3931--3940.

\bibitem{deecap}
Z.~Fei, X.~Yan, S.~Wang, and Q.~Tian, ``Deecap: dynamic early exiting for
  efficient image captioning,'' in \emph{Proceedings of the IEEE/CVF Conference
  on Computer Vision and Pattern Recognition}, 2022, pp. 12\,216--12\,226.

\bibitem{frameexit}
A.~Ghodrati, B.~E. Bejnordi, and A.~Habibian, ``{FrameExit}: Conditional early
  exiting for efficient video recognition,'' in \emph{Proceedings of the
  IEEE/CVF Conference on Computer Vision and Pattern Recognition}, 2021, pp.
  15\,608--15\,618.

\bibitem{stark}
B.~Yan, H.~Peng, J.~Fu, D.~Wang, and H.~Lu, ``Learning spatio-temporal
  transformer for visual tracking,'' in \emph{Proceedings of the IEEE
  International Conference on Computer Vision}, 2021, pp. 10\,448--10\,457.

\bibitem{repvgg}
X.~Ding, X.~Zhang, N.~Ma, J.~Han, G.~Ding, and J.~Sun, ``{RepVGG}: Making
  vgg-style convnets great again,'' in \emph{Proceedings of the IEEE/CVF
  Conference on Computer Vision and Pattern Recognition}, 2021, pp.
  13\,733--13\,742.

\bibitem{transt-m}
X.~Chen, B.~Yan, J.~Zhu, H.~Lu, X.~Ruan, and D.~Wang, ``High-performance
  transformer tracking,'' \emph{IEEE Transactions on Pattern Analysis and
  Machine Intelligence}, vol.~45, no.~7, pp. 8507--8523, 2022.

\bibitem{artrackv2}
Y.~Bai, Z.~Zhao, Y.~Gong, and X.~Wei, ``{ARTrackV2}: Prompting autoregressive
  tracker where to look and how to describe,'' in \emph{Proceedings of the
  IEEE/CVF Conference on Computer Vision and Pattern Recognition}, 2024, pp.
  19\,048--19\,057.

\bibitem{yang2022lavt}
Z.~Yang, J.~Wang, Y.~Tang, K.~Chen, H.~Zhao, and P.~H. Torr, ``{LAVT}:
  Language-aware vision transformer for referring image segmentation,'' in
  \emph{Proceedings of the IEEE/CVF Conference on Computer Vision and Pattern
  Recognition}, 2022, pp. 18\,155--18\,165.

\bibitem{giou}
H.~Rezatofighi, N.~Tsoi, J.~Gwak, A.~Sadeghian, I.~D. Reid, and S.~Savarese,
  ``Generalized intersection over union: {A} metric and a loss for bounding box
  regression,'' in \emph{Proceedings of the IEEE/CVF Conference on Computer
  Vision and Pattern Recognition}, 2019, pp. 658--666.

\bibitem{lasotext}
H.~Fan, H.~Bai, L.~Lin, F.~Yang, P.~Chu, G.~Deng, S.~Yu, Harshit, M.~Huang,
  J.~Liu \emph{et~al.}, ``La{SOT}: A high-quality large-scale single object
  tracking benchmark,'' \emph{International Journal of Computer Vision}, vol.
  129, pp. 439--461, 2021.

\bibitem{got10k}
L.~Huang, X.~Zhao, and K.~Huang, ``{GOT}-10k: {A} large high-diversity
  benchmark for generic object tracking in the wild,'' \emph{IEEE Transactions
  on Pattern Analysis and Machine Intelligence}, vol.~43, no.~5, pp.
  1562--1577, 2019.

\bibitem{trackingnet}
M.~Muller, A.~Bibi, S.~Giancola, S.~Alsubaihi, and B.~Ghanem, ``Tracking{N}et:
  A large-scale dataset and benchmark for object tracking in the wild,'' in
  \emph{European Conference on Computer Vision}, 2018, pp. 300--317.

\bibitem{ocean}
Z.~Zhang, H.~Peng, J.~Fu, B.~Li, and W.~Hu, ``{Ocean}: Object-aware anchor-free
  tracking,'' in \emph{European Conference on Computer Vision}, 2020, pp.
  771--787.

\bibitem{prdimp}
M.~Danelljan, L.~V. Gool, and R.~Timofte, ``Probabilistic regression for visual
  tracking,'' in \emph{Proceedings of the IEEE/CVF Conference on Computer
  Vision and Pattern Recognition}, 2020, pp. 7183--7192.

\bibitem{siamrcnn}
P.~Voigtlaender, J.~Luiten, P.~H.~S. Torr, and B.~Leibe, ``Siam {R-CNN:}
  {V}isual tracking by re-detection,'' in \emph{Proceedings of the IEEE/CVF
  Conference on Computer Vision and Pattern Recognition}, 2020, pp. 6578--6588.

\bibitem{trdimp}
N.~Wang, W.~Zhou, J.~Wang, and H.~Li, ``Transformer meets tracker: Exploiting
  temporal context for robust visual tracking,'' in \emph{Proceedings of the
  IEEE/CVF Conference on Computer Vision and Pattern Recognition}, 2021, pp.
  1571--1580.

\bibitem{cswintt}
Z.~Song, J.~Yu, Y.-P.~P. Chen, and W.~Yang, ``Transformer tracking with cyclic
  shifting window attention,'' in \emph{Proceedings of the IEEE/CVF Conference
  on Computer Vision and Pattern Recognition}, 2022, pp. 8791--8800.

\bibitem{sbt}
F.~Xie, C.~Wang, G.~Wang, Y.~Cao, W.~Yang, and W.~Zeng, ``Correlation-aware
  deep tracking,'' in \emph{Proceedings of the IEEE/CVF Conference on Computer
  Vision and Pattern Recognition}, 2022, pp. 8751--8760.

\bibitem{aiatrack}
S.~Gao, C.~Zhou, C.~Ma, X.~Wang, and J.~Yuan, ``{AiATrack}: Attention in
  attention for transformer visual tracking,'' in \emph{European Conference on
  Computer Vision}.\hskip 1em plus 0.5em minus 0.4em\relax Springer, 2022, pp.
  146--164.

\bibitem{hcat}
X.~Chen, B.~Kang, D.~Wang, D.~Li, and H.~Lu, ``Efficient visual tracking via
  hierarchical cross-attention transformer,'' in \emph{European Conference on
  Computer Vision Workshops}.\hskip 1em plus 0.5em minus 0.4em\relax Springer,
  2023, pp. 461--477.

\bibitem{mat}
H.~Zhao, D.~Wang, and H.~Lu, ``Representation learning for visual object
  tracking by masked appearance transfer,'' in \emph{Proceedings of the
  IEEE/CVF Conference on Computer Vision and Pattern Recognition}, June 2023,
  pp. 18\,696--18\,705.

\bibitem{cttrack}
Z.~Song, R.~Luo, J.~Yu, Y.-P.~P. Chen, and W.~Yang, ``Compact transformer
  tracker with correlative masked modeling,'' in \emph{Proceedings of the AAAI
  Conference on Artificial Intelligence}, 2023, pp. 2321--2329.

\bibitem{coco}
T.-Y. Lin, M.~Maire, S.~J. Belongie, L.~D. Bourdev, R.~B. Girshick, J.~Hays,
  P.~Perona, D.~Ramanan, P.~Doll{\'a}r, and C.~L. Zitnick, ``{Microsoft COCO}:
  {C}ommon objects in context,'' in \emph{European Conference on Computer
  Vision}, 2014, pp. 740--755.

\bibitem{adamw}
I.~Loshchilov and F.~Hutter, ``Decoupled weight decay regularization,'' in
  \emph{International Conference on Learning Representations}, 2018, pp. 1--10.

\bibitem{tnl2k}
X.~Wang, X.~Shu, Z.~Zhang, B.~Jiang, Y.~Wang, Y.~Tian, and F.~Wu, ``Towards
  more flexible and accurate object tracking with natural language: Algorithms
  and benchmark,'' in \emph{Proceedings of the IEEE/CVF Conference on Computer
  Vision and Pattern Recognition}, 2021, pp. 13\,763--13\,773.

\bibitem{nfs}
H.~Kiani~Galoogahi, A.~Fagg, C.~Huang, D.~Ramanan, and S.~Lucey, ``Need for
  speed: {A} benchmark for higher frame rate object tracking,'' in
  \emph{Proceedings of the IEEE International Conference on Computer Vision},
  2017, pp. 1125--1134.

\bibitem{uav}
M.~Mueller, N.~Smith, and B.~Ghanem, ``A benchmark and simulator for {UAV}
  tracking,'' in \emph{European Conference on Computer Vision}, 2016, pp.
  445--461.

\end{thebibliography}

\end{document}